\documentclass{article}
\usepackage{iclr2027_conference,times}
\usepackage{amsmath,amssymb,booktabs,multirow,array,graphicx}
\usepackage{colortbl}
\usepackage{tikz}
\usetikzlibrary{arrows.meta,positioning,fit}
\usepackage{hyperref,url}
\usepackage{enumitem}
\usepackage{placeins}
\definecolor{IFMAPBlue}{RGB}{232,242,252}
\definecolor{MeanGreen}{RGB}{235,247,235}
  
\newcommand{\method}{IF-MAP}
\newcommand{\gnorm}{\ensuremath{G\text{-norm}}}
\newcommand{\iabs}{\ensuremath{I\text{-abs}}}
\newcommand{\meanabs}{\ensuremath{\mathrm{MeanAbs}}}
\newcommand{\SEsel}{\ensuremath{\mathrm{SE}_{\rm sel}}}
\newcommand{\SEtest}{\ensuremath{\mathrm{SE}_{\rm test}}}

\newcommand{\gainse}[2]{\ensuremath{#1\,{\scriptstyle\pm #2}}}
\newif\ifshowresultse
\showresultsefalse
\newcommand{\resultse}[2]{%
  \ifshowresultse
    \shortstack{#1\\{\fontsize{6}{7}\selectfont$\pm #2$}}%
  \else
    #1%
  \fi}
\newcommand{\paircell}[2]{\shortstack{\ensuremath{#1}\\\ensuremath{(#2)}}}

\newif\ifarxivversion
\arxivversiontrue 

\title{IF-MAP: Influence-Function-Inspired Agent Prioritization in Multi-Agent Systems}
\ifarxivversion
  \hypersetup{
    pdftitle={IF-MAP: Influence-Function-Inspired Agent Prioritization in Multi-Agent Systems},
    pdfauthor={Jie Mao, Pingchen Lu, Zikun Qu, Jiahang Sun, Xiangyi Wang, Zhiwei Shang, Zhiyong Wang, Guojun Xiong, Zhongxiang Dai}}
  \author{
    \normalfont\textbf{Jie Mao}\textsuperscript{1}, Pingchen Lu\textsuperscript{1},
    Zikun Qu\textsuperscript{1}, Jiahang Sun\textsuperscript{1},\\
    Xiangyi Wang\textsuperscript{1}, Zhiwei Shang\textsuperscript{1},
    Zhiyong Wang\textsuperscript{2}, Guojun Xiong\textsuperscript{3},
    Zhongxiang Dai\textsuperscript{1,*}\\
    {\normalfont\small \textsuperscript{1}The Chinese University of Hong Kong, Shenzhen}\\
    {\normalfont\small \textsuperscript{2}Harbin Institute of Technology}\\
    {\normalfont\small \textsuperscript{3}School of Engineering and Applied Sciences, Harvard University}\\
    {\normalfont\small \textsuperscript{*}Corresponding author}\\
    {\normalfont\small Code: \url{https://github.com/s1412/IF-MAS}}}
  \iclrfinalcopy
\else
  \hypersetup{
    pdftitle={IF-MAP: Influence-Function-Inspired Agent Prioritization in Multi-Agent Systems},
    pdfauthor={Anonymous authors}}
  \author{Anonymous authors\\Paper under double-blind review}
\fi
\begin{document}
\maketitle
\ifarxivversion
  \lhead{Preprint}
\fi

\begin{abstract}
Improving every agent in an LLM-based multi-agent system can be prohibitively
costly, making it important to identify which agent should be prioritized to
achieve the greatest system-level gain with limited resources. Failure
attribution can inform this decision, but an agent's responsibility for
failures does not directly reveal the system-level gains achievable by
improving it. We therefore introduce \method{}, an influence-function-inspired method
for identifying agents with high potential for system-level improvement. For each agent,
\method{} measures end-to-end loss changes under controlled semantic prompt
perturbations, fits a regularized local response model in a low-dimensional
prompt-embedding subspace, and uses the estimated response norm as a priority
score. This black-box procedure requires no access to model parameters or
gradients and produces agent priorities that can guide subsequent improvement
effort. We evaluate \method{} across six multi-agent workflows spanning code
generation, question answering, and mathematical reasoning, using the same
priorities for prompt optimization and model upgrades. Optimizing the selected
agents' prompts achieves a macro-average gain of 3.81 percentage points,
compared with 1.85 for uniform agent selection. Across five target models,
upgrading the selected agents yields macro-average gains of 7.62--16.42
percentage points, compared with 3.81--8.62 for uniform agent selection.
Despite its probing overhead, \method{} reduces Qwen3-8B and DeepSeek-V3.2
token use by 74.2\% and 35.2\%, respectively, relative to optimizing every
agent, and reduces target-model tokens by 63.6--64.9\% relative to upgrading
every agent.
These results support the use of \method{} to identify promising agents and
direct improvement effort toward greater end-to-end performance gains.
\end{abstract}

\section{Introduction}
Large language model (LLM)-based multi-agent systems distribute complex tasks
across specialized agents that generate solutions, gather evidence, coordinate
reasoning, and verify intermediate or final outputs
\citep{wu2024autogen,zhang2025aflow}. Their performance depends on both the
capabilities of individual agents and the interactions through which their
outputs affect the overall computation. Improving such a system therefore
offers several possible interventions: replacing an agent's model with a more
capable one, fine-tuning its model, refining its instructions, or improving its
tools. However, these interventions need not produce comparable benefits at
different agents. For example, in a question-answering workflow, refining the
answer-generation agent's prompt may improve overall answer accuracy, whereas
refining the prompt of the agent that combines candidate answers may yield
little benefit or even reduce accuracy.
Determining where to direct improvement effort is consequently an important
decision in developing multi-agent systems.

This decision becomes particularly consequential under limited development
compute, evaluation capacity, or deployment resources. Improving all agents may
exceed these resource limits. For example, replacing every agent's model with
a more capable one can substantially increase the inference cost of each
workflow execution, making a system-wide upgrade prohibitively expensive.
We therefore study \emph{agent prioritization}: given a
fixed multi-agent system, select the agent whose improvement is expected to
produce the largest end-to-end performance gain. This formulation captures
the next improvement decision in an ongoing development process. The goal is
to identify a high-value target without carrying out the full improvement
procedure for every candidate agent.

Existing work provides useful foundations for this decision. 
Failure-attribution methods
identify agents and execution steps associated with unsuccessful outcomes,
providing evidence for debugging and system refinement
\citep{zhang2025whowhen}. Nevertheless, responsibility for an observed failure
does not uniquely determine an agent's potential for system-level improvement.
An agent may frequently contribute to errors while offering little room for
improvement, whereas another agent may admit a change that substantially
improves downstream behavior. Intervention-oriented studies further demonstrate
the importance of evaluating actual recovery and realized system-level gains
\citep{ma2026dover,yoon2026diagnosis}. These observations motivate a
complementary objective: estimating which agent is a promising target for
system-level improvement under a limited budget.

Influence functions offer a useful conceptual starting point for this
objective. Classical influence functions characterize how perturbations to
training data affect a learned model and its predictions
\citep{koh2017influence}. We take inspiration from this perturbation-based
perspective and investigate whether controlled changes to an agent can reveal
information useful for prioritizing its improvement. Because an agent's outputs
propagate through the remaining workflow, measuring the resulting end-to-end
loss captures responses beyond that agent's immediate behavior. In the
black-box setting considered here, we obtain this evidence through workflow
executions and semantic representations, without requiring access to model
parameters or gradients. The central hypothesis is that an estimated local
response can reveal agents with substantial improvement potential, providing
a useful signal for allocating improvement effort.

\begin{figure}[t]
\centering
\input{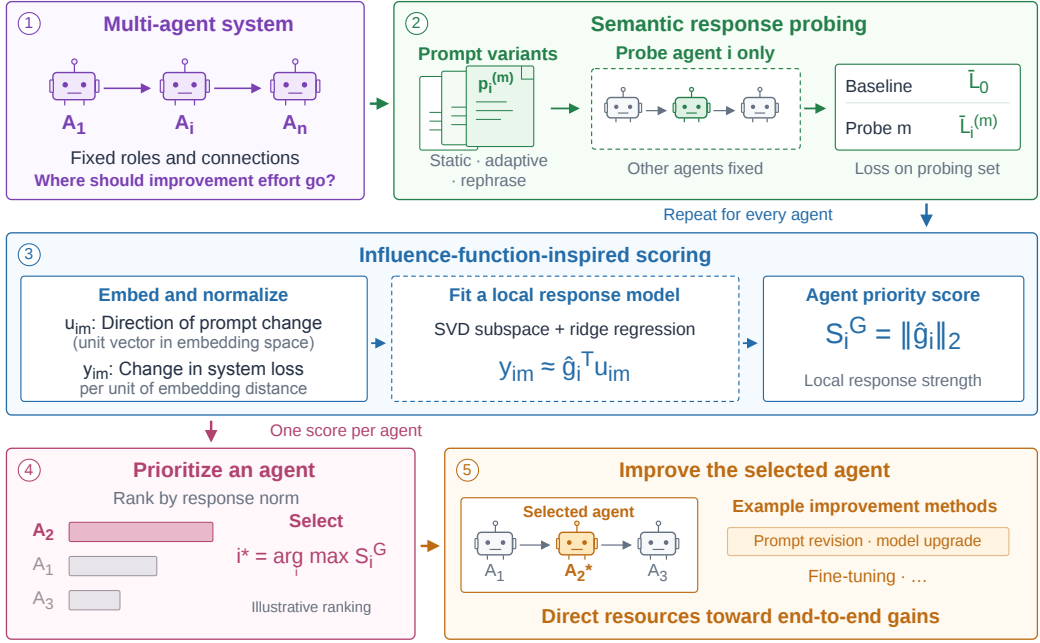}
\vspace{-4mm}
\caption{\textbf{Overview of \method{}.} Given a fixed multi-agent system
\textcircled{\scriptsize 1}, we perturb one agent's prompt at a time and
measure end-to-end losses on the probing set \textcircled{\scriptsize 2}.
Subspace ridge regression on normalized prompt-embedding displacements and
loss responses estimates a local response vector; its norm gives an
influence-function-inspired priority score \textcircled{\scriptsize 3}.
We rank agents by this score \textcircled{\scriptsize 4} and direct improvement
effort toward the top-ranked agent \textcircled{\scriptsize 5}, e.g., through
prompt revision, model upgrades, or fine-tuning. The ranking is illustrative.}
\label{fig:overview}
\end{figure}

We propose \textbf{\method{}}, \emph{Influence Function--Inspired Multi-Agent
Prioritization for System Improvement}. For each agent, \method{} constructs
controlled semantic prompt perturbations and measures their effects on
end-to-end loss while keeping the remaining system configuration fixed. It
represents these perturbations as displacements in a prompt-embedding space
and fits a regularized local response model within a low-dimensional subspace.
The norm of the estimated response vector then provides an agent-priority
score, incorporating both the observed loss changes and the geometry of the
perturbations. Prompt perturbations serve as executable black-box
interventions, and the resulting ranking guides subsequent improvement
effort. This makes \method{} a prioritization step that can be
used alongside existing agent-improvement methods. The score uses estimated
local response sensitivity to define a model-implied local intervention
opportunity, whose relation to realized gains we evaluate through prompt
optimization and model upgrades.

We evaluate \method{} across six task-specific multi-agent
workflows spanning code generation, question answering, and mathematical
reasoning. We assess the usefulness of the same agent priorities through
prompt optimization and upgrades to stronger models. Under prompt
optimization, \method{} achieves a macro-average selected gain of 3.81
percentage points (pp), compared with 1.85 pp for uniform agent selection
(Table~\ref{tab:rq2}). Across five model-upgrade targets, the selected gains
range from 7.62 to 16.42 pp, compared with 3.81--8.62 pp for uniform agent
selection (Table~\ref{tab:rq3}). The selected mean
gains exceed the uniform-selection reference on all six workflows under
prompt optimization and each model-upgrade target. The priorities also align
with observed improvement gains: mean Kendall $\tau_b$ is 0.741 for prompt
optimization and 0.708--0.804 across the five model upgrades. These findings
support \method{} as a useful tool for directing improvement effort toward
agents with substantial potential for system-level gains.

Our contributions are threefold:
\begin{itemize}[leftmargin=*,itemsep=0pt]
\vspace{-2mm}
    \item \textbf{Agent prioritization for system improvement.}
    We formulate the selection of an agent to prioritize under a limited budget
    in a fixed multi-agent system, with expected end-to-end improvement as
    the target quantity.
    \item \textbf{Influence-function-inspired response estimation.}
    We introduce \method{}, a black-box prioritization method that combines
    semantic perturbations, whole-system response measurements, and a
    regularized local model to estimate agent response sensitivity.
    \item \textbf{Empirical validation of agent priorities.}
    We evaluate the same priorities across six workflows through
    prompt optimization and model upgrades, obtaining higher mean selected
    gains than uniform agent selection and measuring ranking agreement with
    observed improvement gains.
\end{itemize}

\begin{figure}[t]
\centering
\makebox[\textwidth][c]{%
  \includegraphics[width=1.0\textwidth]{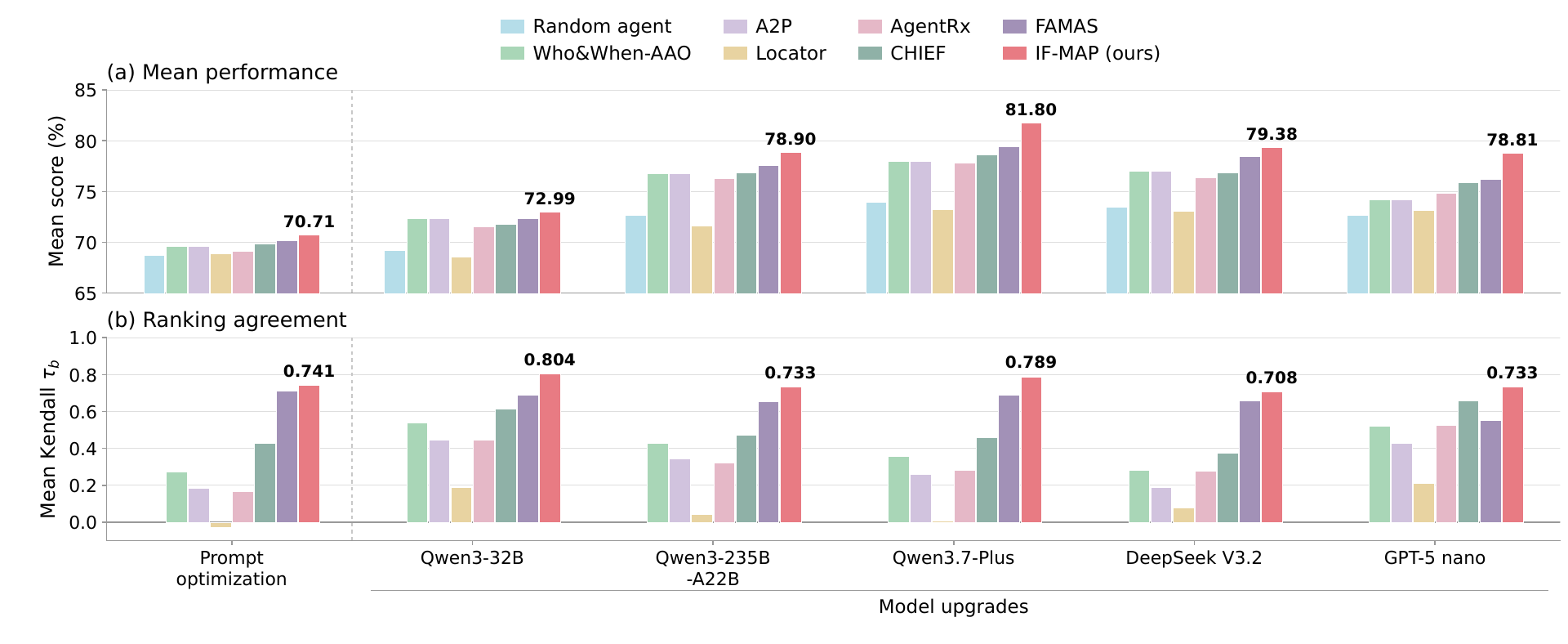}}
\vspace{-10mm}
\caption{Macro-average performance and ranking agreement across six workflows
spanning code generation (HumanEval and MBPP), question answering
(NQ, HotpotQA and DROP), and mathematical problem solving (MATH). 
The base model (before the model upgrade) is Qwen3-8B.
Higher mean performance and ranking agreement (Kendall $\tau_b$) indicate better system performance and stronger agreement between the agent-priority and realized-improvement rankings, respectively.}
\label{fig:macro-performance-kendall}
\end{figure}

\vspace{-2mm}
\section{Agent prioritization in multi-agent systems}
\label{sec:problem}
\vspace{-2mm}
We consider a fixed LLM-based multi-agent system $W$ with $n$ agents,
indexed by $i\in\{1,\ldots,n\}$, prompts $p=(p_1,\ldots,p_n)$, and
assigned models. An agent corresponds to a persistent workflow role and may
execute multiple times within a run. The workflow structure determines how
agents exchange information, so changing one agent can affect both its own
outputs and the behavior of downstream agents. For a task example $x$, the
system-level score $s(W,x)\in[0,1]$ defines the loss
$\ell(W,x)=1-s(W,x)$.

\paragraph{Prioritization objective.}
Under a limited improvement budget, our goal is to identify the agent whose
improvement is expected to yield the largest end-to-end gain. Let
$\mathcal A_i$ denote an improvement procedure applied to agent $i$, with the
workflow structure and the configurations of all other agents fixed. This
procedure may refine the agent's prompt, upgrade or fine-tune its model, or
improve its tools. For a given procedure and budget, the expected utility of
improving agent $i$ is
\begin{equation}
 U_i=\mathbb E_{x,\xi}\left[
 \ell(W,x)-\ell(\mathcal A_i(W;\xi),x)\right],
 \label{eq:utility}
\end{equation}
where the expectation is over task examples $x$ and randomness $\xi$ in
workflow execution and improvement. A prioritization method assigns each agent a
score $S_i$ and selects $\hat\imath\in\arg\max_i S_i$, aiming for a large
$U_{\hat\imath}$ without executing the full improvement procedure for every
agent. \method{} estimates these priorities from development-time response
probes; its ranking can then guide subsequent agent improvement.

\paragraph{Evaluating agent priorities.}
We evaluate the priorities through the gains obtained by improving the
selected agents. Let $\widehat U_i$ be the mean observed test gain from
improving agent $i$. For a unique top-ranked agent, the \emph{selected gain}
is $100\widehat U_{\hat\imath}$ percentage points (pp). If multiple agents
share the highest priority score, we report their average observed improvement
gain, corresponding to selecting one of these agents uniformly at random.
We compare selected gain with no improvement (zero gain),
uniform agent selection ($100n^{-1}\sum_i\widehat U_i$), and the empirical
best agent ($100\max_i\widehat U_i$). Evaluating improvements at every agent
provides an offline reference for assessing the selection; these test outcomes
are not inputs to the prioritization method.

To assess the full ranking, we compute Kendall's $\tau_b$
\citep{kendall1945ties} between the ranking induced by agent-priority scores
and the ranking induced by mean observed test gains from improving each agent.
This metric accounts for ties in both rankings and measures whether agents
assigned higher priority tend to yield larger improvements. Higher values
indicate stronger agreement.
Appendix~\ref{app:statistics} specifies aggregation and uncertainty estimates.

\vspace{-2mm}
\section{IF-MAP: Influence-function-inspired agent prioritization}
\label{sec:method}
\vspace{-2mm}
\method{} is inspired by influence functions, which characterize how
perturbations to training data affect a learned model and its predictions
\citep{koh2017influence}. We adapt this perturbation-response perspective to
agent prioritization: \emph{semantic changes to one agent's prompt reveal how its
behavior affects end-to-end loss}. Our estimator fits a local response model
from black-box workflow executions and prompt embeddings, without computing
classical training-data influences or accessing model parameters, gradients,
or Hessians. As illustrated in Figure~\ref{fig:overview}, we probe each agent,
estimate its local response strength, and use the resulting scores to direct
improvement effort.

\paragraph{Data subsets.}
We use disjoint development subsets for probing, optimization, and validation,
and reserve the test set for frozen interventions; complete data roles and
sizes appear in Appendix~\ref{app:setup}.

\subsection{Probing agents through semantic prompt perturbations}
For each agent $i$, we start from its original prompt $p_i$ and construct at
most ten formal perturbations $p_i^{(m)}$ spanning static, adaptive, and
structure-preserving changes. They retain the agent's role and required output
interface, while structural checks reject unusable or duplicate prompts. The
ten-slot budget is therefore an upper bound on the realized perturbation pool.
Appendix~\ref{app:perturbations} provides the full perturbation protocol.

Each probe changes only agent $i$'s prompt, keeping the other agents' prompts,
all assigned models, and the workflow structure fixed. We execute the full
system on the probing set, so the measured response includes changes propagated through
agent interactions. 
Let $\overline L_0$ denote the mean
baseline loss and $\overline L_i^{(m)}$ the mean loss under perturbation $m$.
The signed effect is $\Delta_{im}=\overline L_0-\overline L_i^{(m)}$, with
positive values indicating improvement.

\subsection{Estimating local response strength}
\label{sec:response-strength}
Our goal is to turn the probe results into a priority score that reflects
how strongly system performance responds to changes in each agent.
Influence functions motivate studying such responses through a local
approximation: they estimate how perturbing training data affects a learned
model and its downstream loss \citep{koh2017influence}. Inspired by this
perspective, we estimate how changes to an agent's prompt affect the
end-to-end loss of the multi-agent system. We learn this response from
observed workflow executions, using prompt embeddings to represent the
changes.

\paragraph{Representing prompt changes and their effects.}
Each probe provides two pieces of information: how the prompt changed and
how the system's loss changed. A fixed embedding model $E$ maps prompts to
L2-normalized vectors, allowing us to represent a prompt change as a
displacement in embedding space. Since perturbations can differ in size,
we normalize both the displacement and the observed loss change by the
embedding distance between the original and perturbed prompts:
\begin{equation}
 v_{im}=E(p_i^{(m)})-E(p_i),\quad
 u_{im}=\frac{v_{im}}{\|v_{im}\|_2},\quad
 y_{im}=\frac{\overline L_i^{(m)}-\overline L_0}{\|v_{im}\|_2}.
 \label{eq:directions}
\end{equation}
Here, $v_{im}$ represents the prompt change, $u_{im}$ describes its
direction, and $y_{im}$ measures the resulting loss change per unit of
embedding distance. Positive values of $y_{im}$ indicate increased loss,
whereas negative values indicate improvement. This normalization compares
responses relative to perturbation size, using embedding distance as a
proxy for that size. We exclude zero or near-zero displacements and let
$M_i$ denote the number of retained measurements.

\paragraph{Combining probes into a stable local response model.}
A single probe measures the response to one particular change. To combine
evidence from several changes, we fit a simple linear model that predicts
each normalized loss response from its direction:
$y_{im}\approx\hat g_i^{\top}u_{im}$.
The vector $\hat g_i$ acts as an estimated local slope: it summarizes which
directions tend to increase or decrease loss and how strongly the system
responds along those directions. This adopts a local perturbation--response
perspective inspired by influence functions, while estimating the response
directly from finite prompt perturbations rather than differentiating through
model training.

There are few probes relative to the embedding dimension, so estimating a
separate response for every embedding coordinate would be unreliable.
We therefore restrict the fit to a low-dimensional space derived from the
observed probe directions. We also use ridge regularization to discourage
large, unstable coefficients. These choices allow the estimate to combine
the available evidence without fitting an unrestricted high-dimensional
response.

Formally, let $V_i$ have rows $u_{im}^{\top}$ and let $y_i$ collect the
responses $y_{im}$. We construct an orthonormal basis $Q_i$ from the
right-singular vectors of the centered direction matrix and set
$Z_i=V_iQ_i$. Centering is used only to construct the basis; the regression
retains the original directions so that they remain paired with their
measured responses. Ridge regression
\citep{hoerl1970ridge}
gives
\begin{equation}
 \hat\beta_i=(Z_i^{\top}Z_i+\lambda I)^{-1}Z_i^{\top}y_i,
 \qquad \hat g_i=Q_i\hat\beta_i,
 \qquad S_i^{G}=\|\hat g_i\|_2.
 \label{eq:ridge}
\end{equation}
The coefficients $\hat\beta_i$ describe the response in the reduced space,
and $\hat g_i$ expresses that response in the original embedding space.
We use Qwen3-Embedding-8B vectors
\citep{zhang2025qwenembedding}, and a default subspace size of
$\max(1,M_i-1)$, bounded by the available SVD dimensions.

\paragraph{Turning response strength into an agent priority.}
For a local displacement $\delta\in\operatorname{span}(Q_i)$, the fitted model
predicts the loss change $\hat g_i^\top\delta$. Under a common local
intervention radius $r$, define the model-implied intervention opportunity as
\begin{equation}
 \widehat U_i(r)
 = \max_{\substack{\delta\in\operatorname{span}(Q_i)\\
                    \|\delta\|_2\le r}}
   -\hat g_i^\top\delta
 = r\|\hat g_i\|_2
 = rS_i^G.
 \label{eq:local-opportunity}
\end{equation}
The equality follows from Cauchy--Schwarz and, for $\hat g_i\ne 0$, is attained
at $\delta_i^*=-r\hat g_i/\|\hat g_i\|_2$. Thus, for a common radius, ranking
agents by $S_i^G$ is equivalent to ranking them by the maximum first-order loss
reduction predicted within the fitted perturbation subspace. Intuitively, a
larger score indicates a steeper local response and a larger predicted
loss-reducing direction. Section~\ref{sec:experiments} evaluates how these
priorities transfer to realizable prompt revisions and model upgrades.

\paragraph{Comparison with response-magnitude scores.}
For ablation, we compare \gnorm{} with \iabs{} and \meanabs{}, which summarize
the largest and average absolute probe effects, respectively. Unlike these
magnitude-only scores, \gnorm{} also uses the directions and sizes of prompt
changes. Appendix~\ref{app:budget} gives the formal definitions.

\subsection{From agent priorities to system improvement}
After ranking agents by $S_i^G$, we apply the improvement procedure only to a
highest-scoring agent while keeping the remaining workflow fixed.

We assess the same priorities through prompt optimization and model upgrades.
For prompt optimization, a shared textual-feedback optimizer revises only the
selected agent using development-set evidence
\citep{pryzant2023protegi,yuksekgonul2025textgrad}. Its full protocol appears
in Appendix~\ref{app:optimizer}.
For model upgrades, the selected agent receives a stronger model while
retaining its original prompt. Final configurations are fixed before
test-set evaluation;
Section~\ref{sec:experiments} details the evaluation protocols and the
model-upgrade results.

\section{Experiments}
\label{sec:experiments}
\subsection{Experimental setup and baselines}
\label{sec:experimental-setup}
\paragraph{Workflows and data.}
We study six AFlow-derived workflows \citep{zhang2025aflow}:
HumanEval and MBPP (code), NQ (factoid question answering), HotpotQA
(multi-hop question answering), DROP (discrete reasoning over passages), and MATH (mathematical problem solving)
\citep{chen2021humaneval,austin2021mbpp,kwiatkowski2019nq,
yang2018hotpotqa,dua2019drop,hendrycks2021math}.
For all six datasets, we start from the benchmark partitions used by AFlow
and divide their development data into the roles specified in
Appendix~\ref{app:setup}. The appendix also specifies sample counts and node
inventories.

\paragraph{Models and configurations.}
Experiments use Qwen3-8B for workflow execution \citep{yang2025qwen3},
DeepSeek-V3.2 for prompt operations \citep{deepseek2025v32}, and five
model-upgrade targets: Qwen3-32B, Qwen3-235B-A22B,
Qwen3.7-Plus \citep{qwen2025instruct2507,alibaba2026qwen37plus},
DeepSeek-V3.2, and GPT-5 nano \citep{openai2026gpt5nano}. DeepSeek's prompt-operation and upgrade
roles are evaluated and accounted for separately. The Qwen3-235B-A22B
target specifically uses the Instruct-2507 checkpoint. Full configurations and
repetitions appear in Appendix~\ref{app:setup}, with uncertainty aggregation
specified in Appendix~\ref{app:statistics}.

\begin{table}[t]
\ifshowresultse
\caption{Prompt optimization with standard errors: complete mean$\pm$SE
version of Table~\ref{tab:rq2}. Each dataset has separate columns for selected
gain and Kendall $\tau_b$; the final pair gives six-dataset macro means.}
\label{tab:rq2-se}
\else
\caption{Prompt optimization: mean selected gain and mean Kendall $\tau_b$
for each dataset and their six-dataset macro means.}
\label{tab:rq2}
\fi
\centering
\fontsize{8}{9}\selectfont
\setlength{\tabcolsep}{1.5pt}
\renewcommand{\arraystretch}{1.18}
\ifshowresultse
\resizebox{\textwidth}{!}{%
\fi
\begin{tabular}{
@{}l
*{6}{cc}
>{\columncolor{MeanGreen}}c
>{\columncolor{MeanGreen}}c
@{}
}
\toprule
Algorithm & \multicolumn{2}{c}{HumanEval} & \multicolumn{2}{c}{MBPP} & \multicolumn{2}{c}{NQ} & \multicolumn{2}{c}{DROP} & \multicolumn{2}{c}{HotpotQA} & \multicolumn{2}{c}{MATH} & \multicolumn{2}{>{\columncolor{MeanGreen}}c}{Macro (6)} \\
\cmidrule(lr){2-3}\cmidrule(lr){4-5}\cmidrule(lr){6-7}\cmidrule(lr){8-9}\cmidrule(lr){10-11}\cmidrule(lr){12-13}\cmidrule(lr){14-15}
 & $\Delta\uparrow$ & $\tau_b\uparrow$ & $\Delta\uparrow$ & $\tau_b\uparrow$ & $\Delta\uparrow$ & $\tau_b\uparrow$ & $\Delta\uparrow$ & $\tau_b\uparrow$ & $\Delta\uparrow$ & $\tau_b\uparrow$ & $\Delta\uparrow$ & $\tau_b\uparrow$ & $\overline\Delta\uparrow$ & $\overline\tau_b\uparrow$ \\
\midrule
\rowcolor{IFMAPBlue}
\textbf{IF-MAP (ours)} & \resultse{+4.44}{0.20} & \resultse{0.556}{0.222} & \resultse{\textbf{+7.91}}{0.00} & \resultse{\textbf{0.889}}{0.111} & \resultse{\textbf{+2.14}}{0.00} & \resultse{\underline{0.778}}{0.222} & \resultse{\textbf{+4.99}}{0.00} & \resultse{\textbf{1.000}}{0.000} & \resultse{\textbf{+1.53}}{0.00} & \resultse{\underline{0.556}}{0.222} & \resultse{\textbf{+1.85}}{0.15} & \resultse{\textbf{0.667}}{0.067} & \cellcolor{MeanGreen}\resultse{\textbf{+3.81}}{0.05} & \cellcolor{MeanGreen}\resultse{\textbf{0.741}}{0.056} \\
Who\&When-AAO & \resultse{+3.23}{1.01} & \resultse{-0.222}{0.401} & \resultse{+3.02}{0.00} & \resultse{0.222}{0.111} & \resultse{\textbf{+2.14}}{0.00} & \resultse{0.333}{0.000} & \resultse{\textbf{+4.99}}{0.00} & \resultse{\textbf{1.000}}{0.000} & \resultse{\textbf{+1.53}}{0.00} & \resultse{0.333}{0.000} & \resultse{+1.38}{0.00} & \resultse{-0.035}{0.070} & \resultse{+2.72}{0.17} & \resultse{0.272}{0.075} \\
A2P & \resultse{+3.23}{1.01} & \resultse{-0.556}{0.222} & \resultse{+3.02}{0.00} & \resultse{0.000}{0.000} & \resultse{\textbf{+2.14}}{0.00} & \resultse{0.333}{0.000} & \resultse{\textbf{+4.99}}{0.00} & \resultse{\textbf{1.000}}{0.000} & \resultse{\textbf{+1.53}}{0.00} & \resultse{0.333}{0.000} & \resultse{+1.38}{0.00} & \resultse{0.004}{0.164} & \resultse{+2.72}{0.17} & \resultse{0.186}{0.028} \\
Adapted locator & \resultse{\underline{+4.55}}{0.00} & \resultse{\textbf{0.816}}{0.000} & \resultse{\underline{+5.27}}{2.64} & \resultse{0.516}{0.289} & \resultse{-0.11}{0.00} & \resultse{-0.333}{0.000} & \resultse{\underline{+2.98}}{2.01} & \resultse{\underline{0.333}}{0.667} & \resultse{-0.55}{0.00} & \resultse{-0.778}{0.222} & \resultse{-0.40}{0.36} & \resultse{-0.713}{0.059} & \resultse{+1.96}{0.46} & \resultse{-0.026}{0.038} \\
AgentRx & \resultse{+3.74}{0.51} & \resultse{-0.161}{0.342} & \resultse{+0.16}{0.16} & \resultse{-0.411}{0.297} & \resultse{\textbf{+2.14}}{0.00} & \resultse{0.333}{0.000} & \resultse{\textbf{+4.99}}{0.00} & \resultse{\textbf{1.000}}{0.000} & \resultse{\textbf{+1.53}}{0.00} & \resultse{0.333}{0.000} & \resultse{+1.05}{0.33} & \resultse{-0.084}{0.221} & \resultse{+2.27}{0.10} & \resultse{0.169}{0.063} \\
CHIEF & \resultse{+4.44}{0.20} & \resultse{0.444}{0.294} & \resultse{+3.80}{2.18} & \resultse{0.365}{0.279} & \resultse{\underline{+1.39}}{0.75} & \resultse{0.111}{0.222} & \resultse{\textbf{+4.99}}{0.00} & \resultse{\textbf{1.000}}{0.000} & \resultse{\textbf{+1.53}}{0.00} & \resultse{0.333}{0.000} & \resultse{\underline{+1.70}}{0.18} & \resultse{\underline{0.305}}{0.058} & \resultse{+2.98}{0.47} & \resultse{0.427}{0.104} \\
FAMAS & \resultse{\textbf{+4.65}}{0.20} & \resultse{\underline{0.717}}{0.199} & \resultse{+4.65}{1.63} & \resultse{\underline{0.669}}{0.122} & \resultse{\textbf{+2.14}}{0.00} & \resultse{\textbf{0.878}}{0.061} & \resultse{\textbf{+4.99}}{0.00} & \resultse{\textbf{1.000}}{0.000} & \resultse{\textbf{+1.53}}{0.00} & \resultse{\textbf{0.778}}{0.222} & \resultse{\textbf{+1.85}}{0.15} & \resultse{0.219}{0.114} & \resultse{\underline{+3.30}}{0.21} & \resultse{\underline{0.710}}{0.067} \\
Uniform-agent reference$^\dagger$ & \resultse{+3.43}{0.40} & \textemdash & \resultse{+2.85}{0.56} & \textemdash & \resultse{+1.34}{0.36} & \textemdash & \resultse{+1.98}{0.18} & \textemdash & \resultse{\underline{+0.33}}{0.36} & \textemdash & \resultse{+1.18}{0.28} & \textemdash & \resultse{+1.85}{0.15} & \textemdash \\
\bottomrule
\end{tabular}
\ifshowresultse
}
\fi
\par\vspace{3pt}
\begin{minipage}{\linewidth}
\fontsize{8}{9}\selectfont
\ifshowresultse
Mean above, $\pm$SE below; $\Delta$ is gain in percentage points.
Method-row SEs summarize three prioritization runs; per-agent intervention
means are computed from five runs. Top ties use uniform tie-averaged gain,
and Kendall retains score ties. $\dagger$: uniform-agent expected gain, shown
as a reference average rather than a selector, with
$\mathrm{SE}_{\rm test}$ across the five intervention runs instead;
\textbf{Bold}/\underline{underline}: largest/second-largest distinct displayed mean per column
within each panel.
\else
Entries report means only; $\Delta$ is gain in percentage points. The complete
mean$\pm$SE results appear in Appendix Table~\ref{tab:rq2-se}. Top ties use
uniform tie-averaged gain, and Kendall retains score ties. $\dagger$ denotes
the uniform-agent expected gain, a reference average rather than a selector;
\textbf{bold}/\underline{underline} marks the
largest/second-largest distinct displayed mean per column.
\fi
\end{minipage}
\end{table}

\paragraph{Baselines.}
We compare \method{} with six failure-attribution, debugging, or weak-agent
localization methods adapted to produce workflow-level agent rankings.
Who\&When-AAO uses the complete conversation log for agent/step failure
attribution \citep{zhang2025whowhen}; A2P uses counterfactual blame reasoning
\citep{west2025a2p}; the adapted locator uses the locator component of Shen
et al. \citep{shen2025locator}; AgentRx uses trajectory-derived constraints
and debugging evidence \citep{barke2026agentrx}; and CHIEF and FAMAS provide
additional failure-attribution procedures
\citep{wang2026chief,ge2026famas}. We use the All-at-Once variant of
Who\&When because the original study reports higher agent-level accuracy when
the full conversation log is available. We adapt all six methods to produce
workflow-level agent rankings from development trajectories.
Appendix~\ref{app:baselines} describes the common evaluation setting,
aggregation procedure, and method-specific adaptations.

\subsection{Main Results}
\paragraph{Prompt optimization.}
Table~\ref{tab:rq2} shows that \method{} achieves a macro-average selected gain
of $3.81$ pp, which is $1.96$ pp above the uniform-agent reference and $0.51$
pp above FAMAS, the strongest external selector on this measure. Its selected
gain exceeds the uniform-agent reference on all six workflows, with the
largest margins on MBPP and DROP ($5.06$ and $3.01$ pp, respectively), so the
aggregate improvement is not confined to one task family. \method{} also has
the highest mean Kendall $\tau_b$ ($0.741$ versus $0.710$ for FAMAS), showing
that it achieves a better overall ordering of agents by their realized
prompt-improvement gains, not only a better top-agent selection.

\ifshowresultse
\begin{table}[t]
\ifshowresultse
\caption{Model upgrades with standard errors: complete mean$\pm$SE version of
Table~\ref{tab:rq3}. Each panel uses one target model and its own
uniform-agent reference.}
\label{tab:rq3-se}
\else
\caption{Model upgrades: selecting an agent for upgrade. Each panel uses one target
model and its own uniform-agent reference (the $\dagger$ row).
Mean gain and mean Kendall are reported for each dataset and in macro average.}
\label{tab:rq3}
\fi

\centering
\fontsize{8}{9}\selectfont
\setlength{\tabcolsep}{1.3pt}
\renewcommand{\arraystretch}{1.18}

\resizebox{\textwidth}{!}{%
\begin{tabular}{
@{}l
*{6}{cc}
>{\columncolor{MeanGreen}}c
>{\columncolor{MeanGreen}}c
@{}
}
\toprule

Algorithm
& \multicolumn{2}{c}{HumanEval}
& \multicolumn{2}{c}{MBPP}
& \multicolumn{2}{c}{NQ}
& \multicolumn{2}{c}{DROP}
& \multicolumn{2}{c}{HotpotQA}
& \multicolumn{2}{c}{MATH}
& \multicolumn{2}{>{\columncolor{MeanGreen}}c}{Macro (6)}
\\

\cmidrule(lr){2-3}
\cmidrule(lr){4-5}
\cmidrule(lr){6-7}
\cmidrule(lr){8-9}
\cmidrule(lr){10-11}
\cmidrule(lr){12-13}
\cmidrule(lr){14-15}

& $\Delta\uparrow$ & $\tau_b\uparrow$
& $\Delta\uparrow$ & $\tau_b\uparrow$
& $\Delta\uparrow$ & $\tau_b\uparrow$
& $\Delta\uparrow$ & $\tau_b\uparrow$
& $\Delta\uparrow$ & $\tau_b\uparrow$
& $\Delta\uparrow$ & $\tau_b\uparrow$
& $\overline\Delta\uparrow$ & $\overline\tau_b\uparrow$
\\

\midrule

\multicolumn{15}{@{}l}{\textbf{A. Qwen3-32B}} \\
\addlinespace[2pt]

\rowcolor{IFMAPBlue}
\textbf{IF-MAP (ours)}
& \resultse{\underline{+11.52}}{3.64}
& \resultse{\underline{0.778}}{0.222}
& \resultse{\textbf{+7.44}}{0.00}
& \resultse{\textbf{0.889}}{0.111}
& \resultse{\textbf{+10.63}}{0.00}
& \resultse{\underline{0.778}}{0.222}
& \resultse{\textbf{+6.63}}{0.00}
& \resultse{\textbf{1.000}}{0.000}
& \resultse{\textbf{+5.74}}{0.00}
& \resultse{\underline{0.778}}{0.222}
& \resultse{\textbf{+3.76}}{1.19}
& \resultse{\textbf{0.600}}{0.115}
& \cellcolor{MeanGreen}\resultse{\textbf{+7.62}}{0.54}
& \cellcolor{MeanGreen}\resultse{\textbf{0.804}}{0.036}
\\

Who\&When-AAO
& \resultse{+10.30}{4.85}
& \resultse{0.494}{0.417}
& \resultse{+6.74}{0.00}
& \resultse{0.222}{0.111}
& \resultse{\textbf{+10.63}}{0.00}
& \resultse{0.333}{0.000}
& \resultse{\textbf{+6.63}}{0.00}
& \resultse{\textbf{1.000}}{0.000}
& \resultse{\textbf{+5.74}}{0.00}
& \resultse{\textbf{1.000}}{0.000}
& \resultse{+1.65}{0.00}
& \resultse{0.176}{0.070}
& \resultse{\underline{+6.95}}{0.81}
& \resultse{0.538}{0.074}
\\

A2P
& \resultse{+10.30}{4.85}
& \resultse{0.111}{0.222}
& \resultse{+6.74}{0.00}
& \resultse{0.000}{0.000}
& \resultse{\textbf{+10.63}}{0.00}
& \resultse{0.333}{0.000}
& \resultse{\textbf{+6.63}}{0.00}
& \resultse{\textbf{1.000}}{0.000}
& \resultse{\textbf{+5.74}}{0.00}
& \resultse{\textbf{1.000}}{0.000}
& \resultse{+1.65}{0.00}
& \resultse{0.219}{0.167}
& \resultse{\underline{+6.95}}{0.81}
& \resultse{0.444}{0.027}
\\

Adapted locator
& \resultse{+9.70}{0.00}
& \resultse{\textbf{0.816}}{0.000}
& \resultse{+4.88}{2.56}
& \resultse{0.516}{0.289}
& \resultse{-0.30}{0.00}
& \resultse{-0.333}{0.000}
& \resultse{\underline{+5.56}}{1.07}
& \resultse{\underline{0.333}}{0.667}
& \resultse{+0.40}{0.00}
& \resultse{-0.111}{0.222}
& \resultse{-0.91}{0.51}
& \resultse{-0.102}{0.058}
& \resultse{+3.22}{0.44}
& \resultse{0.187}{0.040}
\\

AgentRx
& \resultse{\textbf{+12.73}}{2.42}
& \resultse{0.605}{0.307}
& \resultse{+0.16}{0.39}
& \resultse{-0.411}{0.297}
& \resultse{\textbf{+10.63}}{0.00}
& \resultse{0.333}{0.000}
& \resultse{\textbf{+6.63}}{0.00}
& \resultse{\textbf{1.000}}{0.000}
& \resultse{\textbf{+5.74}}{0.00}
& \resultse{\textbf{1.000}}{0.000}
& \resultse{+1.09}{0.56}
& \resultse{0.141}{0.217}
& \resultse{+6.16}{0.41}
& \resultse{0.445}{0.066}
\\

CHIEF
& \resultse{\underline{+11.52}}{3.64}
& \resultse{0.717}{0.199}
& \resultse{+5.04}{2.06}
& \resultse{0.365}{0.279}
& \resultse{\underline{+6.99}}{3.65}
& \resultse{0.111}{0.222}
& \resultse{\textbf{+6.63}}{0.00}
& \resultse{\textbf{1.000}}{0.000}
& \resultse{\textbf{+5.74}}{0.00}
& \resultse{\textbf{1.000}}{0.000}
& \resultse{\underline{+2.39}}{1.32}
& \resultse{0.502}{0.263}
& \resultse{+6.38}{0.62}
& \resultse{0.616}{0.026}
\\

FAMAS
& \resultse{+7.88}{3.64}
& \resultse{0.444}{0.294}
& \resultse{\underline{+6.98}}{0.23}
& \resultse{\underline{0.669}}{0.122}
& \resultse{\textbf{+10.63}}{0.00}
& \resultse{\textbf{0.878}}{0.061}
& \resultse{\textbf{+6.63}}{0.00}
& \resultse{\textbf{1.000}}{0.000}
& \resultse{\textbf{+5.74}}{0.00}
& \resultse{0.556}{0.222}
& \resultse{\textbf{+3.76}}{1.19}
& \resultse{\underline{0.579}}{0.169}
& \resultse{+6.94}{0.45}
& \resultse{\underline{0.688}}{0.024}
\\

Uniform-agent reference$^\dagger$
& \resultse{+6.67}{0.69}
& \textemdash
& \resultse{+3.72}{0.12}
& \textemdash
& \resultse{+3.88}{0.27}
& \textemdash
& \resultse{+5.03}{0.89}
& \textemdash
& \resultse{\underline{+2.17}}{0.21}
& \textemdash
& \resultse{+1.42}{1.76}
& \textemdash
& \resultse{+3.81}{0.35}
& \textemdash
\\

\midrule

\multicolumn{15}{@{}l}{\textbf{B. Qwen3-235B-A22B}} \\
\addlinespace[2pt]

\rowcolor{IFMAPBlue}
\textbf{IF-MAP (ours)}
& \resultse{\underline{+19.19}}{5.05}
& \resultse{\underline{0.778}}{0.222}
& \resultse{\textbf{+11.63}}{0.00}
& \resultse{\textbf{0.889}}{0.111}
& \resultse{\textbf{+21.78}}{0.00}
& \resultse{\underline{0.778}}{0.222}
& \resultse{\textbf{+11.13}}{0.00}
& \resultse{\textbf{1.000}}{0.000}
& \resultse{\textbf{+7.88}}{0.00}
& \resultse{\underline{0.556}}{0.222}
& \resultse{\textbf{+9.55}}{3.95}
& \resultse{\textbf{0.400}}{0.115}
& \cellcolor{MeanGreen}\resultse{\textbf{+13.53}}{0.77}
& \cellcolor{MeanGreen}\resultse{\textbf{0.733}}{0.040}
\\

Who\&When-AAO
& \resultse{+16.16}{8.08}
& \resultse{0.494}{0.417}
& \resultse{+7.21}{0.00}
& \resultse{0.222}{0.111}
& \resultse{\textbf{+21.78}}{0.00}
& \resultse{0.333}{0.000}
& \resultse{\textbf{+11.13}}{0.00}
& \resultse{\textbf{1.000}}{0.000}
& \resultse{\textbf{+7.88}}{0.00}
& \resultse{0.333}{0.000}
& \resultse{+4.41}{0.00}
& \resultse{0.176}{0.141}
& \resultse{+11.43}{1.35}
& \resultse{0.426}{0.087}
\\

A2P
& \resultse{+16.16}{8.08}
& \resultse{0.111}{0.222}
& \resultse{+7.21}{0.00}
& \resultse{0.000}{0.000}
& \resultse{\textbf{+21.78}}{0.00}
& \resultse{0.333}{0.000}
& \resultse{\textbf{+11.13}}{0.00}
& \resultse{\textbf{1.000}}{0.000}
& \resultse{\textbf{+7.88}}{0.00}
& \resultse{0.333}{0.000}
& \resultse{+4.41}{0.00}
& \resultse{0.290}{0.100}
& \resultse{+11.43}{1.35}
& \resultse{0.345}{0.026}
\\

Adapted locator
& \resultse{+16.67}{0.00}
& \resultse{\textbf{0.816}}{0.000}
& \resultse{+7.60}{4.03}
& \resultse{0.516}{0.289}
& \resultse{-0.98}{0.00}
& \resultse{-0.333}{0.000}
& \resultse{\underline{+9.87}}{1.26}
& \resultse{\underline{0.333}}{0.667}
& \resultse{+0.53}{0.00}
& \resultse{-0.778}{0.222}
& \resultse{+3.72}{0.14}
& \resultse{-0.305}{0.058}
& \resultse{+6.23}{0.61}
& \resultse{0.042}{0.039}
\\

AgentRx
& \resultse{\textbf{+20.20}}{4.04}
& \resultse{0.605}{0.307}
& \resultse{+0.85}{1.32}
& \resultse{-0.411}{0.297}
& \resultse{\textbf{+21.78}}{0.00}
& \resultse{0.333}{0.000}
& \resultse{\textbf{+11.13}}{0.00}
& \resultse{\textbf{1.000}}{0.000}
& \resultse{\textbf{+7.88}}{0.00}
& \resultse{0.333}{0.000}
& \resultse{+3.81}{0.60}
& \resultse{0.072}{0.374}
& \resultse{+10.94}{0.78}
& \resultse{0.322}{0.091}
\\

CHIEF
& \resultse{\underline{+19.19}}{5.05}
& \resultse{0.717}{0.199}
& \resultse{+7.44}{2.35}
& \resultse{0.365}{0.279}
& \resultse{\underline{+14.19}}{7.59}
& \resultse{0.111}{0.222}
& \resultse{\textbf{+11.13}}{0.00}
& \resultse{\textbf{1.000}}{0.000}
& \resultse{\textbf{+7.88}}{0.00}
& \resultse{0.333}{0.000}
& \resultse{\underline{+9.00}}{2.62}
& \resultse{\underline{0.305}}{0.058}
& \resultse{+11.47}{1.09}
& \resultse{0.472}{0.037}
\\

FAMAS
& \resultse{+14.14}{5.05}
& \resultse{0.444}{0.294}
& \resultse{\underline{+8.68}}{1.47}
& \resultse{\underline{0.669}}{0.122}
& \resultse{\textbf{+21.78}}{0.00}
& \resultse{\textbf{0.878}}{0.061}
& \resultse{\textbf{+11.13}}{0.00}
& \resultse{\textbf{1.000}}{0.000}
& \resultse{\textbf{+7.88}}{0.00}
& \resultse{\textbf{0.778}}{0.222}
& \resultse{\textbf{+9.55}}{3.95}
& \resultse{0.145}{0.247}
& \resultse{\underline{+12.19}}{0.43}
& \resultse{\underline{0.652}}{0.059}
\\

Uniform-agent reference$^\dagger$
& \resultse{+11.11}{0.00}
& \textemdash
& \resultse{+5.47}{0.07}
& \textemdash
& \resultse{+7.81}{0.22}
& \textemdash
& \resultse{+9.24}{0.50}
& \textemdash
& \resultse{\underline{+3.55}}{0.21}
& \textemdash
& \resultse{+6.45}{0.66}
& \textemdash
& \resultse{+7.27}{0.15}
& \textemdash
\\

\midrule

\multicolumn{15}{@{}l}{\textbf{C. Qwen3.7-Plus}} \\
\addlinespace[2pt]

\rowcolor{IFMAPBlue}
\textbf{IF-MAP (ours)}
& \resultse{\underline{+21.21}}{6.06}
& \resultse{\underline{0.778}}{0.222}
& \resultse{\textbf{+21.17}}{0.00}
& \resultse{\textbf{0.889}}{0.111}
& \resultse{\textbf{+19.78}}{0.00}
& \resultse{\underline{0.778}}{0.222}
& \resultse{\textbf{+12.82}}{0.00}
& \resultse{\textbf{1.000}}{0.000}
& \resultse{\textbf{+10.46}}{0.00}
& \resultse{\underline{0.556}}{0.222}
& \resultse{\textbf{+13.09}}{0.66}
& \resultse{\textbf{0.733}}{0.176}
& \cellcolor{MeanGreen}\resultse{\textbf{+16.42}}{0.96}
& \cellcolor{MeanGreen}\resultse{\textbf{0.789}}{0.068}
\\

Who\&When-AAO
& \resultse{+18.18}{9.09}
& \resultse{0.494}{0.417}
& \resultse{+9.07}{0.00}
& \resultse{0.222}{0.111}
& \resultse{\textbf{+19.78}}{0.00}
& \resultse{0.333}{0.000}
& \resultse{\textbf{+12.82}}{0.00}
& \resultse{\textbf{1.000}}{0.000}
& \resultse{\textbf{+10.46}}{0.00}
& \resultse{0.333}{0.000}
& \resultse{+5.23}{0.00}
& \resultse{-0.246}{0.070}
& \resultse{+12.59}{1.52}
& \resultse{0.356}{0.074}
\\

A2P
& \resultse{+18.18}{9.09}
& \resultse{0.111}{0.222}
& \resultse{+9.07}{0.00}
& \resultse{0.000}{0.000}
& \resultse{\textbf{+19.78}}{0.00}
& \resultse{0.333}{0.000}
& \resultse{\textbf{+12.82}}{0.00}
& \resultse{\textbf{1.000}}{0.000}
& \resultse{\textbf{+10.46}}{0.00}
& \resultse{0.333}{0.000}
& \resultse{+5.23}{0.00}
& \resultse{-0.211}{0.161}
& \resultse{+12.59}{1.52}
& \resultse{0.261}{0.028}
\\

Adapted locator
& \resultse{+18.18}{0.00}
& \resultse{\textbf{0.816}}{0.000}
& \resultse{\underline{+14.03}}{7.13}
& \resultse{0.516}{0.289}
& \resultse{-0.12}{0.00}
& \resultse{-0.333}{0.000}
& \resultse{\underline{+11.03}}{1.79}
& \resultse{\underline{0.333}}{0.667}
& \resultse{+0.14}{0.00}
& \resultse{-0.778}{0.222}
& \resultse{+3.86}{0.28}
& \resultse{-0.509}{0.058}
& \resultse{+7.85}{1.10}
& \resultse{0.008}{0.039}
\\

AgentRx
& \resultse{\textbf{+22.73}}{4.55}
& \resultse{0.605}{0.307}
& \resultse{+2.70}{2.93}
& \resultse{-0.411}{0.297}
& \resultse{\textbf{+19.78}}{0.00}
& \resultse{0.333}{0.000}
& \resultse{\textbf{+12.82}}{0.00}
& \resultse{\textbf{1.000}}{0.000}
& \resultse{\textbf{+10.46}}{0.00}
& \resultse{0.333}{0.000}
& \resultse{+6.05}{0.81}
& \resultse{-0.159}{0.138}
& \resultse{+12.42}{1.11}
& \resultse{0.284}{0.060}
\\

CHIEF
& \resultse{\underline{+21.21}}{6.06}
& \resultse{0.717}{0.199}
& \resultse{+12.93}{4.12}
& \resultse{0.365}{0.279}
& \resultse{\underline{+13.15}}{6.63}
& \resultse{0.111}{0.222}
& \resultse{\textbf{+12.82}}{0.00}
& \resultse{\textbf{1.000}}{0.000}
& \resultse{\textbf{+10.46}}{0.00}
& \resultse{0.333}{0.000}
& \resultse{\underline{+9.08}}{2.49}
& \resultse{0.235}{0.185}
& \resultse{+13.28}{0.72}
& \resultse{0.460}{0.015}
\\

FAMAS
& \resultse{+15.15}{6.06}
& \resultse{0.444}{0.294}
& \resultse{+13.10}{4.03}
& \resultse{\underline{0.669}}{0.122}
& \resultse{\textbf{+19.78}}{0.00}
& \resultse{\textbf{0.878}}{0.061}
& \resultse{\textbf{+12.82}}{0.00}
& \resultse{\textbf{1.000}}{0.000}
& \resultse{\textbf{+10.46}}{0.00}
& \resultse{\textbf{0.778}}{0.222}
& \resultse{\textbf{+13.09}}{0.66}
& \resultse{\underline{0.364}}{0.165}
& \resultse{\underline{+14.07}}{1.57}
& \resultse{\underline{0.689}}{0.113}
\\

Uniform-agent reference$^\dagger$
& \resultse{+12.12}{0.00}
& \textemdash
& \resultse{+9.64}{0.20}
& \textemdash
& \resultse{+6.99}{0.23}
& \textemdash
& \resultse{+10.13}{0.19}
& \textemdash
& \resultse{\underline{+4.31}}{0.13}
& \textemdash
& \resultse{+8.52}{0.50}
& \textemdash
& \resultse{+8.62}{0.11}
& \textemdash
\\

\bottomrule
\end{tabular}%
}

\par\vspace{3pt}

\begin{minipage}{\linewidth}
\fontsize{8}{9}\selectfont
\ifshowresultse
Mean above, $\pm$SE below. Method-row SEs summarize three prioritization runs;
uniform-agent reference rows use $\mathrm{SE}_{\rm test}$ across five intervention
runs and are reference averages rather than selectors. Other notation and
formatting conventions follow Appendix
Table~\ref{tab:rq2-se}.
\else
Entries report means only; the complete mean$\pm$SE results appear in Appendix
Table~\ref{tab:rq3-se}. Uniform-agent rows are reference averages rather than
selectors. Other notation and formatting conventions follow
Table~\ref{tab:rq2}.
\fi
\end{minipage}
\end{table}
\begin{table}[t]
\ifshowresultse
\caption{Model upgrades with standard errors: complete mean$\pm$SE version of
Table~\ref{tab:rq3}. Each panel uses one target model and its own
uniform-agent reference.}
\label{tab:rq3-se-new-models}
\else
\caption{Model upgrades: selecting an agent for upgrade. Each panel uses one target
model and its own uniform-agent reference (the $\dagger$ row).
Mean gain and mean Kendall are reported for each dataset and in macro average.}
\label{tab:rq3}
\fi

\centering
\fontsize{8}{9}\selectfont
\setlength{\tabcolsep}{1.3pt}
\renewcommand{\arraystretch}{1.18}

\resizebox{\textwidth}{!}{%
\begin{tabular}{
@{}l
*{6}{cc}
>{\columncolor{MeanGreen}}c
>{\columncolor{MeanGreen}}c
@{}
}
\toprule

Algorithm
& \multicolumn{2}{c}{HumanEval}
& \multicolumn{2}{c}{MBPP}
& \multicolumn{2}{c}{NQ}
& \multicolumn{2}{c}{DROP}
& \multicolumn{2}{c}{HotpotQA}
& \multicolumn{2}{c}{MATH}
& \multicolumn{2}{>{\columncolor{MeanGreen}}c}{Macro (6)}
\\

\cmidrule(lr){2-3}
\cmidrule(lr){4-5}
\cmidrule(lr){6-7}
\cmidrule(lr){8-9}
\cmidrule(lr){10-11}
\cmidrule(lr){12-13}
\cmidrule(lr){14-15}

& $\Delta\uparrow$ & $\tau_b\uparrow$
& $\Delta\uparrow$ & $\tau_b\uparrow$
& $\Delta\uparrow$ & $\tau_b\uparrow$
& $\Delta\uparrow$ & $\tau_b\uparrow$
& $\Delta\uparrow$ & $\tau_b\uparrow$
& $\Delta\uparrow$ & $\tau_b\uparrow$
& $\overline\Delta\uparrow$ & $\overline\tau_b\uparrow$
\\

\midrule

\multicolumn{15}{@{}l}{\textbf{D. DeepSeek-V3.2}} \\
\addlinespace[2pt]
\rowcolor{IFMAPBlue}
\textbf{IF-MAP (ours)}
& \resultse{\textbf{+15.76}}{3.03}
& \resultse{\underline{0.778}}{0.222}
& \resultse{\textbf{+12.33}}{0.00}
& \resultse{\textbf{0.889}}{0.111}
& \resultse{\textbf{+23.08}}{0.00}
& \resultse{\underline{0.778}}{0.222}
& \resultse{\textbf{+8.97}}{0.00}
& \resultse{\textbf{1.000}}{0.000}
& \resultse{\textbf{+5.18}}{0.00}
& \resultse{\underline{0.556}}{0.222}
& \resultse{\textbf{+9.59}}{2.70}
& \resultse{\textbf{0.246}}{0.186}
& \cellcolor{MeanGreen}\resultse{\textbf{+12.48}}{0.48}
& \cellcolor{MeanGreen}\resultse{\textbf{0.708}}{0.069} \\
Who\&When-AAO
& \resultse{+12.53}{6.26}
& \resultse{0.494}{0.417}
& \resultse{+8.60}{0.00}
& \resultse{0.222}{0.111}
& \resultse{\textbf{+23.08}}{0.00}
& \resultse{0.333}{0.000}
& \resultse{\textbf{+8.97}}{0.00}
& \resultse{\textbf{1.000}}{0.000}
& \resultse{\textbf{+5.18}}{0.00}
& \resultse{0.333}{0.000}
& \resultse{+2.20}{0.00}
& \resultse{-0.704}{0.098}
& \cellcolor{MeanGreen}\resultse{+10.09}{1.04}
& \cellcolor{MeanGreen}\resultse{0.280}{0.081} \\
A2P
& \resultse{+12.53}{6.26}
& \resultse{0.111}{0.222}
& \resultse{+8.60}{0.00}
& \resultse{0.000}{0.000}
& \resultse{\textbf{+23.08}}{0.00}
& \resultse{0.333}{0.000}
& \resultse{\textbf{+8.97}}{0.00}
& \resultse{\textbf{1.000}}{0.000}
& \resultse{\textbf{+5.18}}{0.00}
& \resultse{0.333}{0.000}
& \resultse{+2.20}{0.00}
& \resultse{-0.639}{0.127}
& \cellcolor{MeanGreen}\resultse{+10.09}{1.04}
& \cellcolor{MeanGreen}\resultse{0.190}{0.028} \\
Adapted locator
& \resultse{+14.24}{0.00}
& \resultse{\textbf{0.816}}{0.000}
& \resultse{+8.29}{4.03}
& \resultse{0.516}{0.289}
& \resultse{+1.70}{0.00}
& \resultse{-0.333}{0.000}
& \resultse{\underline{+7.63}}{1.34}
& \resultse{\underline{0.333}}{0.667}
& \resultse{+1.47}{0.00}
& \resultse{-0.778}{0.222}
& \resultse{+3.86}{0.33}
& \resultse{-0.074}{0.096}
& \cellcolor{MeanGreen}\resultse{+6.20}{0.55}
& \cellcolor{MeanGreen}\resultse{0.080}{0.028} \\
AgentRx
& \resultse{\underline{+15.66}}{3.13}
& \resultse{0.605}{0.307}
& \resultse{+0.93}{0.70}
& \resultse{-0.411}{0.297}
& \resultse{\textbf{+23.08}}{0.00}
& \resultse{0.333}{0.000}
& \resultse{\textbf{+8.97}}{0.00}
& \resultse{\textbf{1.000}}{0.000}
& \resultse{\textbf{+5.18}}{0.00}
& \resultse{0.333}{0.000}
& \resultse{+2.87}{0.66}
& \resultse{-0.201}{0.224}
& \cellcolor{MeanGreen}\resultse{+9.45}{0.64}
& \cellcolor{MeanGreen}\resultse{0.277}{0.086} \\
CHIEF
& \resultse{\textbf{+15.76}}{3.03}
& \resultse{0.717}{0.199}
& \resultse{+7.75}{2.92}
& \resultse{0.365}{0.279}
& \resultse{\underline{+15.96}}{7.13}
& \resultse{0.111}{0.222}
& \resultse{\textbf{+8.97}}{0.00}
& \resultse{\textbf{1.000}}{0.000}
& \resultse{\textbf{+5.18}}{0.00}
& \resultse{0.333}{0.000}
& \resultse{\underline{+6.01}}{3.16}
& \resultse{-0.289}{0.198}
& \cellcolor{MeanGreen}\resultse{+9.94}{0.92}
& \cellcolor{MeanGreen}\resultse{0.373}{0.014} \\
FAMAS
& \resultse{+12.73}{3.03}
& \resultse{0.444}{0.294}
& \resultse{\underline{+9.84}}{1.24}
& \resultse{\underline{0.669}}{0.122}
& \resultse{\textbf{+23.08}}{0.00}
& \resultse{\textbf{0.878}}{0.061}
& \resultse{\textbf{+8.97}}{0.00}
& \resultse{\textbf{1.000}}{0.000}
& \resultse{\textbf{+5.18}}{0.00}
& \resultse{\textbf{0.778}}{0.222}
& \resultse{\textbf{+9.59}}{2.70}
& \resultse{\underline{0.192}}{0.103}
& \cellcolor{MeanGreen}\resultse{\underline{+11.57}}{0.26}
& \cellcolor{MeanGreen}\resultse{\underline{0.660}}{0.090} \\
Uniform-agent reference$^\dagger$
& \resultse{+9.49}{0.49}
& \textemdash
& \resultse{+5.87}{0.29}
& \textemdash
& \resultse{+8.89}{0.18}
& \textemdash
& \resultse{+6.96}{0.15}
& \textemdash
& \resultse{\underline{+3.01}}{0.19}
& \textemdash
& \resultse{+5.28}{0.38}
& \textemdash
& \cellcolor{MeanGreen}\resultse{+6.58}{0.18}
& \textemdash \\
\midrule
\multicolumn{15}{@{}l}{\textbf{E. GPT-5 nano}} \\
\addlinespace[2pt]
\rowcolor{IFMAPBlue}
\textbf{IF-MAP (ours)}
& \resultse{\underline{+19.39}}{5.45}
& \resultse{\underline{0.778}}{0.222}
& \resultse{\textbf{+17.21}}{0.00}
& \resultse{\textbf{0.556}}{0.111}
& \resultse{\textbf{+14.10}}{0.00}
& \resultse{0.556}{0.222}
& \resultse{\textbf{+6.50}}{0.00}
& \resultse{\textbf{1.000}}{0.000}
& \resultse{\textbf{+3.03}}{0.00}
& \resultse{\underline{0.778}}{0.222}
& \resultse{\textbf{+11.24}}{1.87}
& \resultse{\textbf{0.733}}{0.176}
& \cellcolor{MeanGreen}\resultse{\textbf{+11.91}}{0.80}
& \cellcolor{MeanGreen}\resultse{\textbf{0.733}}{0.106} \\
Who\&When-AAO
& \resultse{+16.77}{8.08}
& \resultse{0.494}{0.417}
& \resultse{+2.09}{0.00}
& \resultse{-0.111}{0.111}
& \resultse{\textbf{+14.10}}{0.00}
& \resultse{\textbf{1.000}}{0.000}
& \resultse{\textbf{+6.50}}{0.00}
& \resultse{\textbf{1.000}}{0.000}
& \resultse{\textbf{+3.03}}{0.00}
& \resultse{\textbf{1.000}}{0.000}
& \resultse{+1.38}{0.00}
& \resultse{-0.246}{0.070}
& \cellcolor{MeanGreen}\resultse{+7.31}{1.35}
& \cellcolor{MeanGreen}\resultse{0.523}{0.074} \\
A2P
& \resultse{+16.77}{8.08}
& \resultse{0.111}{0.222}
& \resultse{+2.09}{0.00}
& \resultse{-0.333}{0.000}
& \resultse{\textbf{+14.10}}{0.00}
& \resultse{\textbf{1.000}}{0.000}
& \resultse{\textbf{+6.50}}{0.00}
& \resultse{\textbf{1.000}}{0.000}
& \resultse{\textbf{+3.03}}{0.00}
& \resultse{\textbf{1.000}}{0.000}
& \resultse{+1.38}{0.00}
& \resultse{-0.211}{0.161}
& \cellcolor{MeanGreen}\resultse{+7.31}{1.35}
& \cellcolor{MeanGreen}\resultse{0.428}{0.028} \\
Adapted locator
& \resultse{+16.67}{0.00}
& \resultse{\textbf{0.816}}{0.000}
& \resultse{\underline{+11.32}}{5.89}
& \resultse{\underline{0.415}}{0.381}
& \resultse{+2.14}{0.00}
& \resultse{0.333}{0.000}
& \resultse{\underline{+5.22}}{1.28}
& \resultse{\underline{0.333}}{0.667}
& \resultse{+1.38}{0.00}
& \resultse{-0.111}{0.222}
& \resultse{+0.69}{0.14}
& \resultse{-0.509}{0.058}
& \cellcolor{MeanGreen}\resultse{+6.24}{0.91}
& \cellcolor{MeanGreen}\resultse{0.213}{0.077} \\
AgentRx
& \resultse{\textbf{+20.81}}{4.04}
& \resultse{0.605}{0.307}
& \resultse{+1.01}{1.47}
& \resultse{-0.289}{0.418}
& \resultse{\textbf{+14.10}}{0.00}
& \resultse{\textbf{1.000}}{0.000}
& \resultse{\textbf{+6.50}}{0.00}
& \resultse{\textbf{1.000}}{0.000}
& \resultse{\textbf{+3.03}}{0.00}
& \resultse{\textbf{1.000}}{0.000}
& \resultse{+2.26}{0.88}
& \resultse{-0.159}{0.138}
& \cellcolor{MeanGreen}\resultse{+7.95}{0.87}
& \cellcolor{MeanGreen}\resultse{0.526}{0.078} \\
CHIEF
& \resultse{\underline{+19.39}}{5.45}
& \resultse{0.717}{0.199}
& \resultse{+7.75}{4.76}
& \resultse{0.233}{0.217}
& \resultse{\underline{+10.11}}{3.99}
& \resultse{\underline{0.778}}{0.222}
& \resultse{\textbf{+6.50}}{0.00}
& \resultse{\textbf{1.000}}{0.000}
& \resultse{\textbf{+3.03}}{0.00}
& \resultse{\textbf{1.000}}{0.000}
& \resultse{\underline{+7.11}}{3.39}
& \resultse{0.235}{0.185}
& \cellcolor{MeanGreen}\resultse{+8.98}{0.32}
& \cellcolor{MeanGreen}\resultse{\underline{0.660}}{0.026} \\
FAMAS
& \resultse{+13.94}{5.45}
& \resultse{0.444}{0.294}
& \resultse{+7.13}{5.04}
& \resultse{0.304}{0.122}
& \resultse{\textbf{+14.10}}{0.00}
& \resultse{0.655}{0.161}
& \resultse{\textbf{+6.50}}{0.00}
& \resultse{\textbf{1.000}}{0.000}
& \resultse{\textbf{+3.03}}{0.00}
& \resultse{0.556}{0.222}
& \resultse{\textbf{+11.24}}{1.87}
& \resultse{\underline{0.364}}{0.165}
& \cellcolor{MeanGreen}\resultse{\underline{+9.32}}{1.44}
& \cellcolor{MeanGreen}\resultse{0.554}{0.059} \\
Uniform-agent reference$^\dagger$
& \resultse{+11.31}{0.45}
& \textemdash
& \resultse{+5.70}{0.55}
& \textemdash
& \resultse{+5.86}{0.28}
& \textemdash
& \resultse{+4.58}{0.36}
& \textemdash
& \resultse{\underline{+1.51}}{0.33}
& \textemdash
& \resultse{+5.87}{0.66}
& \textemdash
& \cellcolor{MeanGreen}\resultse{+5.81}{0.19}
& \textemdash \\
\bottomrule
\end{tabular}%
}

\par\vspace{3pt}

\begin{minipage}{\linewidth}
\fontsize{8}{9}\selectfont
\ifshowresultse
Mean above, $\pm$SE below. Method-row SEs summarize three prioritization runs;
uniform-agent reference rows use $\mathrm{SE}_{\rm test}$ across five intervention
runs and are reference averages rather than selectors. Other notation and
formatting conventions follow Appendix
Table~\ref{tab:rq2-se}.
\else
Entries report means only; the complete mean$\pm$SE results appear in Appendix
Table~\ref{tab:rq3-se}. Uniform-agent rows are reference averages rather than
selectors. Other notation and formatting conventions follow
Table~\ref{tab:rq2}.
\fi
\end{minipage}
\end{table}
\else
\begin{table}[t]
\ifshowresultse
\caption{Model upgrades with standard errors: complete mean$\pm$SE version of
Table~\ref{tab:rq3}. Each panel uses one target model and its own
uniform-agent reference.}
\label{tab:rq3-se}
\else
\caption{Model upgrades: selecting an agent for upgrade. Panel A averages the
metrics over Qwen3-32B, Qwen3-235B-A22B, and Qwen3.7-Plus; panels B and C
report DeepSeek-V3.2 and GPT-5 nano separately. Each panel includes its
uniform-agent reference (the $\dagger$ row).
Mean gain and mean Kendall are reported for each dataset and in macro average.}
\label{tab:rq3}
\fi

\centering
\fontsize{8}{9}\selectfont
\setlength{\tabcolsep}{1.3pt}
\renewcommand{\arraystretch}{1.18}

\resizebox{\textwidth}{!}{%
\begin{tabular}{
@{}l
*{6}{cc}
>{\columncolor{MeanGreen}}c
>{\columncolor{MeanGreen}}c
@{}
}
\toprule

Algorithm
& \multicolumn{2}{c}{HumanEval}
& \multicolumn{2}{c}{MBPP}
& \multicolumn{2}{c}{NQ}
& \multicolumn{2}{c}{DROP}
& \multicolumn{2}{c}{HotpotQA}
& \multicolumn{2}{c}{MATH}
& \multicolumn{2}{>{\columncolor{MeanGreen}}c}{Macro (6)}
\\

\cmidrule(lr){2-3}
\cmidrule(lr){4-5}
\cmidrule(lr){6-7}
\cmidrule(lr){8-9}
\cmidrule(lr){10-11}
\cmidrule(lr){12-13}
\cmidrule(lr){14-15}

& $\Delta\uparrow$ & $\tau_b\uparrow$
& $\Delta\uparrow$ & $\tau_b\uparrow$
& $\Delta\uparrow$ & $\tau_b\uparrow$
& $\Delta\uparrow$ & $\tau_b\uparrow$
& $\Delta\uparrow$ & $\tau_b\uparrow$
& $\Delta\uparrow$ & $\tau_b\uparrow$
& $\overline\Delta\uparrow$ & $\overline\tau_b\uparrow$
\\

\midrule

\multicolumn{15}{@{}l}{\textbf{A. Qwen3-32B / Qwen3-235B-A22B / Qwen3.7-Plus (mean)}} \\
\addlinespace[2pt]
\rowcolor{IFMAPBlue}
\textbf{IF-MAP (ours)}
& \resultse{\underline{+17.31}}{4.92}
& \resultse{\underline{0.778}}{0.222}
& \resultse{\textbf{+13.41}}{0.00}
& \resultse{\textbf{0.889}}{0.111}
& \resultse{\textbf{+17.40}}{0.00}
& \resultse{\underline{0.778}}{0.222}
& \resultse{\textbf{+10.19}}{0.00}
& \resultse{\textbf{1.000}}{0.000}
& \resultse{\textbf{+8.03}}{0.00}
& \resultse{\underline{0.630}}{0.074}
& \resultse{\textbf{+8.80}}{1.93}
& \resultse{\textbf{0.578}}{0.059}
& \cellcolor{MeanGreen}\resultse{\textbf{+12.52}}{0.71}
& \cellcolor{MeanGreen}\resultse{\textbf{0.775}}{0.022} \\
Who\&When-AAO
& \resultse{+14.88}{7.34}
& \resultse{0.494}{0.417}
& \resultse{+7.67}{0.00}
& \resultse{0.222}{0.111}
& \resultse{\textbf{+17.40}}{0.00}
& \resultse{0.333}{0.000}
& \resultse{\textbf{+10.19}}{0.00}
& \resultse{\textbf{1.000}}{0.000}
& \resultse{\textbf{+8.03}}{0.00}
& \resultse{0.556}{0.000}
& \resultse{+3.76}{0.00}
& \resultse{0.035}{0.081}
& \cellcolor{MeanGreen}\resultse{+10.32}{1.22}
& \cellcolor{MeanGreen}\resultse{0.440}{0.078} \\
A2P
& \resultse{+14.88}{7.34}
& \resultse{0.111}{0.222}
& \resultse{+7.67}{0.00}
& \resultse{0.000}{0.000}
& \resultse{\textbf{+17.40}}{0.00}
& \resultse{0.333}{0.000}
& \resultse{\textbf{+10.19}}{0.00}
& \resultse{\textbf{1.000}}{0.000}
& \resultse{\textbf{+8.03}}{0.00}
& \resultse{0.556}{0.000}
& \resultse{+3.76}{0.00}
& \resultse{0.099}{0.142}
& \cellcolor{MeanGreen}\resultse{+10.32}{1.22}
& \cellcolor{MeanGreen}\resultse{0.350}{0.027} \\
Adapted locator
& \resultse{+14.85}{0.00}
& \resultse{\textbf{0.816}}{0.000}
& \resultse{+8.84}{4.57}
& \resultse{0.516}{0.289}
& \resultse{-0.47}{0.00}
& \resultse{-0.333}{0.000}
& \resultse{\underline{+8.82}}{1.37}
& \resultse{\underline{0.333}}{0.667}
& \resultse{+0.36}{0.00}
& \resultse{-0.556}{0.222}
& \resultse{+2.22}{0.31}
& \resultse{-0.305}{0.058}
& \cellcolor{MeanGreen}\resultse{+5.77}{0.72}
& \cellcolor{MeanGreen}\resultse{0.079}{0.039} \\
AgentRx
& \resultse{\textbf{+18.55}}{3.67}
& \resultse{0.605}{0.307}
& \resultse{+1.23}{1.54}
& \resultse{-0.411}{0.297}
& \resultse{\textbf{+17.40}}{0.00}
& \resultse{0.333}{0.000}
& \resultse{\textbf{+10.19}}{0.00}
& \resultse{\textbf{1.000}}{0.000}
& \resultse{\textbf{+8.03}}{0.00}
& \resultse{0.556}{0.000}
& \resultse{+3.65}{0.11}
& \resultse{0.018}{0.219}
& \cellcolor{MeanGreen}\resultse{+9.84}{0.77}
& \cellcolor{MeanGreen}\resultse{0.350}{0.070} \\
CHIEF
& \resultse{\underline{+17.31}}{4.92}
& \resultse{0.717}{0.199}
& \resultse{+8.47}{2.65}
& \resultse{0.365}{0.279}
& \resultse{\underline{+11.44}}{5.96}
& \resultse{0.111}{0.222}
& \resultse{\textbf{+10.19}}{0.00}
& \resultse{\textbf{1.000}}{0.000}
& \resultse{\textbf{+8.03}}{0.00}
& \resultse{0.556}{0.000}
& \resultse{\underline{+6.82}}{2.06}
& \resultse{0.347}{0.160}
& \cellcolor{MeanGreen}\resultse{+10.38}{0.81}
& \cellcolor{MeanGreen}\resultse{0.516}{0.022} \\
FAMAS
& \resultse{+12.39}{4.92}
& \resultse{0.444}{0.294}
& \resultse{\underline{+9.59}}{1.91}
& \resultse{\underline{0.669}}{0.122}
& \resultse{\textbf{+17.40}}{0.00}
& \resultse{\textbf{0.878}}{0.061}
& \resultse{\textbf{+10.19}}{0.00}
& \resultse{\textbf{1.000}}{0.000}
& \resultse{\textbf{+8.03}}{0.00}
& \resultse{\textbf{0.704}}{0.074}
& \resultse{\textbf{+8.80}}{1.93}
& \resultse{\underline{0.363}}{0.142}
& \cellcolor{MeanGreen}\resultse{\underline{+11.07}}{0.82}
& \cellcolor{MeanGreen}\resultse{\underline{0.676}}{0.065} \\
Uniform-agent reference$^\dagger$
& \resultse{+9.97}{0.00}
& \textemdash
& \resultse{+6.28}{0.00}
& \textemdash
& \resultse{+6.23}{0.00}
& \textemdash
& \resultse{+8.13}{0.00}
& \textemdash
& \resultse{\underline{+3.34}}{0.00}
& \textemdash
& \resultse{+5.46}{0.00}
& \textemdash
& \cellcolor{MeanGreen}\resultse{+6.57}{0.00}
& \textemdash \\
\midrule
\multicolumn{15}{@{}l}{\textbf{B. DeepSeek-V3.2}} \\
\addlinespace[2pt]
\rowcolor{IFMAPBlue}
\textbf{IF-MAP (ours)}
& \resultse{\textbf{+15.76}}{3.03}
& \resultse{\underline{0.778}}{0.222}
& \resultse{\textbf{+12.33}}{0.00}
& \resultse{\textbf{0.889}}{0.111}
& \resultse{\textbf{+23.08}}{0.00}
& \resultse{\underline{0.778}}{0.222}
& \resultse{\textbf{+8.97}}{0.00}
& \resultse{\textbf{1.000}}{0.000}
& \resultse{\textbf{+5.18}}{0.00}
& \resultse{\underline{0.556}}{0.222}
& \resultse{\textbf{+9.59}}{2.70}
& \resultse{\textbf{0.246}}{0.186}
& \cellcolor{MeanGreen}\resultse{\textbf{+12.48}}{0.48}
& \cellcolor{MeanGreen}\resultse{\textbf{0.708}}{0.069} \\
Who\&When-AAO
& \resultse{+12.53}{6.26}
& \resultse{0.494}{0.417}
& \resultse{+8.60}{0.00}
& \resultse{0.222}{0.111}
& \resultse{\textbf{+23.08}}{0.00}
& \resultse{0.333}{0.000}
& \resultse{\textbf{+8.97}}{0.00}
& \resultse{\textbf{1.000}}{0.000}
& \resultse{\textbf{+5.18}}{0.00}
& \resultse{0.333}{0.000}
& \resultse{+2.20}{0.00}
& \resultse{-0.704}{0.098}
& \cellcolor{MeanGreen}\resultse{+10.09}{1.04}
& \cellcolor{MeanGreen}\resultse{0.280}{0.081} \\
A2P
& \resultse{+12.53}{6.26}
& \resultse{0.111}{0.222}
& \resultse{+8.60}{0.00}
& \resultse{0.000}{0.000}
& \resultse{\textbf{+23.08}}{0.00}
& \resultse{0.333}{0.000}
& \resultse{\textbf{+8.97}}{0.00}
& \resultse{\textbf{1.000}}{0.000}
& \resultse{\textbf{+5.18}}{0.00}
& \resultse{0.333}{0.000}
& \resultse{+2.20}{0.00}
& \resultse{-0.639}{0.127}
& \cellcolor{MeanGreen}\resultse{+10.09}{1.04}
& \cellcolor{MeanGreen}\resultse{0.190}{0.028} \\
Adapted locator
& \resultse{+14.24}{0.00}
& \resultse{\textbf{0.816}}{0.000}
& \resultse{+8.29}{4.03}
& \resultse{0.516}{0.289}
& \resultse{+1.70}{0.00}
& \resultse{-0.333}{0.000}
& \resultse{\underline{+7.63}}{1.34}
& \resultse{\underline{0.333}}{0.667}
& \resultse{+1.47}{0.00}
& \resultse{-0.778}{0.222}
& \resultse{+3.86}{0.33}
& \resultse{-0.074}{0.096}
& \cellcolor{MeanGreen}\resultse{+6.20}{0.55}
& \cellcolor{MeanGreen}\resultse{0.080}{0.028} \\
AgentRx
& \resultse{\underline{+15.66}}{3.13}
& \resultse{0.605}{0.307}
& \resultse{+0.93}{0.70}
& \resultse{-0.411}{0.297}
& \resultse{\textbf{+23.08}}{0.00}
& \resultse{0.333}{0.000}
& \resultse{\textbf{+8.97}}{0.00}
& \resultse{\textbf{1.000}}{0.000}
& \resultse{\textbf{+5.18}}{0.00}
& \resultse{0.333}{0.000}
& \resultse{+2.87}{0.66}
& \resultse{-0.201}{0.224}
& \cellcolor{MeanGreen}\resultse{+9.45}{0.64}
& \cellcolor{MeanGreen}\resultse{0.277}{0.086} \\
CHIEF
& \resultse{\textbf{+15.76}}{3.03}
& \resultse{0.717}{0.199}
& \resultse{+7.75}{2.92}
& \resultse{0.365}{0.279}
& \resultse{\underline{+15.96}}{7.13}
& \resultse{0.111}{0.222}
& \resultse{\textbf{+8.97}}{0.00}
& \resultse{\textbf{1.000}}{0.000}
& \resultse{\textbf{+5.18}}{0.00}
& \resultse{0.333}{0.000}
& \resultse{\underline{+6.01}}{3.16}
& \resultse{-0.289}{0.198}
& \cellcolor{MeanGreen}\resultse{+9.94}{0.92}
& \cellcolor{MeanGreen}\resultse{0.373}{0.014} \\
FAMAS
& \resultse{+12.73}{3.03}
& \resultse{0.444}{0.294}
& \resultse{\underline{+9.84}}{1.24}
& \resultse{\underline{0.669}}{0.122}
& \resultse{\textbf{+23.08}}{0.00}
& \resultse{\textbf{0.878}}{0.061}
& \resultse{\textbf{+8.97}}{0.00}
& \resultse{\textbf{1.000}}{0.000}
& \resultse{\textbf{+5.18}}{0.00}
& \resultse{\textbf{0.778}}{0.222}
& \resultse{\textbf{+9.59}}{2.70}
& \resultse{\underline{0.192}}{0.103}
& \cellcolor{MeanGreen}\resultse{\underline{+11.57}}{0.26}
& \cellcolor{MeanGreen}\resultse{\underline{0.660}}{0.090} \\
Uniform-agent reference$^\dagger$
& \resultse{+9.49}{0.49}
& \textemdash
& \resultse{+5.87}{0.29}
& \textemdash
& \resultse{+8.89}{0.18}
& \textemdash
& \resultse{+6.96}{0.15}
& \textemdash
& \resultse{\underline{+3.01}}{0.19}
& \textemdash
& \resultse{+5.28}{0.38}
& \textemdash
& \cellcolor{MeanGreen}\resultse{+6.58}{0.18}
& \textemdash \\
\midrule
\multicolumn{15}{@{}l}{\textbf{C. GPT-5 nano}} \\
\addlinespace[2pt]
\rowcolor{IFMAPBlue}
\textbf{IF-MAP (ours)}
& \resultse{\underline{+19.39}}{5.45}
& \resultse{\underline{0.778}}{0.222}
& \resultse{\textbf{+17.21}}{0.00}
& \resultse{\textbf{0.556}}{0.111}
& \resultse{\textbf{+14.10}}{0.00}
& \resultse{0.556}{0.222}
& \resultse{\textbf{+6.50}}{0.00}
& \resultse{\textbf{1.000}}{0.000}
& \resultse{\textbf{+3.03}}{0.00}
& \resultse{\underline{0.778}}{0.222}
& \resultse{\textbf{+11.24}}{1.87}
& \resultse{\textbf{0.733}}{0.176}
& \cellcolor{MeanGreen}\resultse{\textbf{+11.91}}{0.80}
& \cellcolor{MeanGreen}\resultse{\textbf{0.733}}{0.106} \\
Who\&When-AAO
& \resultse{+16.77}{8.08}
& \resultse{0.494}{0.417}
& \resultse{+2.09}{0.00}
& \resultse{-0.111}{0.111}
& \resultse{\textbf{+14.10}}{0.00}
& \resultse{\textbf{1.000}}{0.000}
& \resultse{\textbf{+6.50}}{0.00}
& \resultse{\textbf{1.000}}{0.000}
& \resultse{\textbf{+3.03}}{0.00}
& \resultse{\textbf{1.000}}{0.000}
& \resultse{+1.38}{0.00}
& \resultse{-0.246}{0.070}
& \cellcolor{MeanGreen}\resultse{+7.31}{1.35}
& \cellcolor{MeanGreen}\resultse{0.523}{0.074} \\
A2P
& \resultse{+16.77}{8.08}
& \resultse{0.111}{0.222}
& \resultse{+2.09}{0.00}
& \resultse{-0.333}{0.000}
& \resultse{\textbf{+14.10}}{0.00}
& \resultse{\textbf{1.000}}{0.000}
& \resultse{\textbf{+6.50}}{0.00}
& \resultse{\textbf{1.000}}{0.000}
& \resultse{\textbf{+3.03}}{0.00}
& \resultse{\textbf{1.000}}{0.000}
& \resultse{+1.38}{0.00}
& \resultse{-0.211}{0.161}
& \cellcolor{MeanGreen}\resultse{+7.31}{1.35}
& \cellcolor{MeanGreen}\resultse{0.428}{0.028} \\
Adapted locator
& \resultse{+16.67}{0.00}
& \resultse{\textbf{0.816}}{0.000}
& \resultse{\underline{+11.32}}{5.89}
& \resultse{\underline{0.415}}{0.381}
& \resultse{+2.14}{0.00}
& \resultse{0.333}{0.000}
& \resultse{\underline{+5.22}}{1.28}
& \resultse{\underline{0.333}}{0.667}
& \resultse{+1.38}{0.00}
& \resultse{-0.111}{0.222}
& \resultse{+0.69}{0.14}
& \resultse{-0.509}{0.058}
& \cellcolor{MeanGreen}\resultse{+6.24}{0.91}
& \cellcolor{MeanGreen}\resultse{0.213}{0.077} \\
AgentRx
& \resultse{\textbf{+20.81}}{4.04}
& \resultse{0.605}{0.307}
& \resultse{+1.01}{1.47}
& \resultse{-0.289}{0.418}
& \resultse{\textbf{+14.10}}{0.00}
& \resultse{\textbf{1.000}}{0.000}
& \resultse{\textbf{+6.50}}{0.00}
& \resultse{\textbf{1.000}}{0.000}
& \resultse{\textbf{+3.03}}{0.00}
& \resultse{\textbf{1.000}}{0.000}
& \resultse{+2.26}{0.88}
& \resultse{-0.159}{0.138}
& \cellcolor{MeanGreen}\resultse{+7.95}{0.87}
& \cellcolor{MeanGreen}\resultse{0.526}{0.078} \\
CHIEF
& \resultse{\underline{+19.39}}{5.45}
& \resultse{0.717}{0.199}
& \resultse{+7.75}{4.76}
& \resultse{0.233}{0.217}
& \resultse{\underline{+10.11}}{3.99}
& \resultse{\underline{0.778}}{0.222}
& \resultse{\textbf{+6.50}}{0.00}
& \resultse{\textbf{1.000}}{0.000}
& \resultse{\textbf{+3.03}}{0.00}
& \resultse{\textbf{1.000}}{0.000}
& \resultse{\underline{+7.11}}{3.39}
& \resultse{0.235}{0.185}
& \cellcolor{MeanGreen}\resultse{+8.98}{0.32}
& \cellcolor{MeanGreen}\resultse{\underline{0.660}}{0.026} \\
FAMAS
& \resultse{+13.94}{5.45}
& \resultse{0.444}{0.294}
& \resultse{+7.13}{5.04}
& \resultse{0.304}{0.122}
& \resultse{\textbf{+14.10}}{0.00}
& \resultse{0.655}{0.161}
& \resultse{\textbf{+6.50}}{0.00}
& \resultse{\textbf{1.000}}{0.000}
& \resultse{\textbf{+3.03}}{0.00}
& \resultse{0.556}{0.222}
& \resultse{\textbf{+11.24}}{1.87}
& \resultse{\underline{0.364}}{0.165}
& \cellcolor{MeanGreen}\resultse{\underline{+9.32}}{1.44}
& \cellcolor{MeanGreen}\resultse{0.554}{0.059} \\
Uniform-agent reference$^\dagger$
& \resultse{+11.31}{0.45}
& \textemdash
& \resultse{+5.70}{0.55}
& \textemdash
& \resultse{+5.86}{0.28}
& \textemdash
& \resultse{+4.58}{0.36}
& \textemdash
& \resultse{\underline{+1.51}}{0.33}
& \textemdash
& \resultse{+5.87}{0.66}
& \textemdash
& \cellcolor{MeanGreen}\resultse{+5.81}{0.19}
& \textemdash \\
\bottomrule
\end{tabular}%
}

\par\vspace{3pt}

\begin{minipage}{\linewidth}
\fontsize{8}{9}\selectfont
\ifshowresultse
Mean above, $\pm$SE below. Method-row SEs summarize three prioritization runs;
uniform-agent reference rows use $\mathrm{SE}_{\rm test}$ across five intervention
runs and are reference averages rather than selectors. Other notation and
formatting conventions follow Appendix
Table~\ref{tab:rq2-se}.
\else
Entries report means only; the complete mean$\pm$SE results appear in Appendix
Table~\ref{tab:rq3-se}. Uniform-agent rows are reference averages rather than
selectors. Other notation and formatting conventions follow
Table~\ref{tab:rq2}.
\fi
\end{minipage}
\end{table}
\fi

\paragraph{Model upgrades.}
The prompt-derived priorities are reused without target-specific
reprioritization. Table~\ref{tab:rq3} gives
macro-average selected gains of $12.52$, $12.48$, and $11.91$ pp for
the mean across the three Qwen upgrade targets, DeepSeek-V3.2, and GPT-5 nano,
respectively.
The corresponding uniform-agent references are $6.57$, $6.58$, and $5.81$ pp,
and the mean Kendall $\tau_b$ values for \method{} are $0.775$, $0.708$, and
$0.733$. Target-wise uncertainty, per-node gains, and complete rankings appear
in Appendix~\ref{app:nodegains}.
Across the five individual targets, selecting with \method{} yields
$1.86$--$2.05$ times the gain of upgrading a uniformly chosen agent.
IF-MAP also exceeds the strongest external selector baseline by $1.45$,
$0.91$, and $2.59$ pp in the three displayed panels and has the highest
macro-average ranking agreement in each panel.
The same priorities are used across all five target models, and
their consistent performance suggests that the probes identify agent-level
leverage that transfers across replacement models. The uniform-agent reference averages
all agents within each dataset, so the gap over that row measures the benefit
of deciding \emph{where} to upgrade rather than the capability gain of the
target model alone.


\subsection{Cost analysis}
Semantic probing evaluates each perturbation through the complete workflow,
requiring more Qwen3-8B executions than most attribution baselines. It uses
only 3.6--34.1\% of their DeepSeek-V3.2 requests and 1.5--16.6\% of their
DeepSeek tokens, shifting computation from diagnostic-model calls to
base-model workflow execution.

The same effect is more consequential for downstream improvement. Compared
with applying the same prompt optimizer to every agent, selecting one agent
with \method{} uses
30.69\% of the Qwen requests, 25.82\% of the Qwen tokens, 57.16\% of the
DeepSeek requests, and 64.78\% of the DeepSeek tokens. For model replacement,
FullSwap (i.e., upgrading every agent to the target model) avoids the
Qwen3-8B and DeepSeek selection overhead, but invokes the
target model for every agent: it uses 2.61--3.16 times as
many target-model requests and 2.75--2.85 times as many target-model tokens as
\method{}. Thus, the practical benefit of prioritization is not simply fewer
total calls, but concentrating optimization and target-model calls on a
single promising agent while using a cheaper model to decide where that effort
should be spent. Complete request and token counts are reported in Appendix
Tables~\ref{tab:prioritization-cost}--\ref{tab:model-upgrade-cost}.

\subsection{Ablation Study}
\begin{table}[t]
\caption{Macro-average ablation results over six datasets.
\method{} uses G-norm and the full probe pool
(at most 10 probes per node).
Scoring variants replace only the scoring rule;
budget variants retain G-norm.
Model upgrades use per-node gains averaged over all five target models.
Dataset means are equally weighted.}
\label{tab:abl-macro}
\centering
\small
\setlength{\tabcolsep}{4pt}
\begin{tabular}{lrrrr}
\toprule
 & \multicolumn{2}{c}{Prompt optimization}
 & \multicolumn{2}{c}{Model upgrades} \\
\cmidrule(lr){2-3}\cmidrule(lr){4-5}
Method & Gain (pp) $\uparrow$ & $\tau_b\uparrow$
       & Gain (pp) $\uparrow$ & $\tau_b\uparrow$ \\
\midrule
\textbf{IF-MAP}
 & $+3.812$ & $0.741$ & $+12.394$ & $0.778$ \\
\multicolumn{5}{l}{\textit{Scoring-rule ablation}} \\
w/ I-abs
 & $+3.276$ & $0.752$ & $+9.396$ & $0.641$ \\
w/ MeanAbs
 & $+3.576$ & $0.774$ & $+9.090$ & $0.663$ \\
\addlinespace[3pt]
\multicolumn{5}{l}{\textit{Probe-budget comparison}} \\
w/ 3 probes
 & $+2.452$ & $0.289$ & $+6.744$ & $0.178$ \\
w/ 5 probes
 & $+2.903$ & $0.415$ & $+7.806$ & $0.304$ \\
\bottomrule
\end{tabular}
\end{table}

Table~\ref{tab:abl-macro} reports two matched ablations, macro-averaged with
equal weight across the six workflows. The scoring-rule variants use the same
probes and realized intervention gains, changing only how probe responses are
converted into priorities. The budget variants all use \gnorm{} but retain
three, five, or up to ten formal probes per agent (adopted by our main
\method{} configuration). For model upgrades, each
agent's gain is averaged across all five target models before aggregation.
Complete aggregation rules appear in Appendix~\ref{app:statistics}.

\paragraph{Scoring-rule ablation.}
This ablation directly tests the Section~\ref{sec:response-strength} choice to
fit a geometry-aware local response rather than use magnitude-only summaries.
Using identical probes and test outcomes, \gnorm{} obtains the highest
selected gains: 3.812 pp for prompt optimization and 12.394 pp for model
upgrades, exceeding \iabs{} by 0.536 and 2.998 pp and \meanabs{} by 0.236 and
3.304 pp, respectively. The larger model-upgrade advantage suggests that the
direction and displacement information used by \gnorm{} transfers beyond
prompt editing. However, \meanabs{} has slightly higher prompt-ranking
agreement ($0.774$ versus $0.741$) while yielding lower selected gain,
showing that agreement over the full ranking need not identify the best
top-ranked intervention target.

\paragraph{Probe-budget comparison.}
With \gnorm{}, increasing the budget from three to five probes raises selected
gain from 2.452 to 2.903 pp for prompt optimization and from 6.744 to 7.806 pp
for model upgrades, while mean Kendall $\tau_b$ increases by $0.126$ for both.
The full pool of up to ten probes further raises the gains to 3.812 and 12.394
pp and substantially improves ranking agreement, suggesting that five probes
provide insufficient aggregate directional coverage. Per-workflow effects are
not monotonic, and the reduced variants retain auxiliary adaptive evaluations;
the comparison therefore supports a larger formal response pool rather than a
proportional cost reduction.

\section{Related work}
\paragraph{Attribution, debugging, and improvement.}
Failure-attribution and debugging methods localize agents or steps associated
with unsuccessful outcomes through full-trajectory attribution,
counterfactual blame, constraint-based evidence, and other localization
signals
\citep{zhang2025whowhen,west2025a2p,shen2025locator,barke2026agentrx,
wang2026chief,ge2026famas}. We adapt six such methods as external selectors in
Section~\ref{sec:experimental-setup}. DoVer emphasizes that attribution should
be validated through realized interventions
\citep{ma2026dover}. CICA shows more directly that, in its experimental
setting, the module assigned the greatest responsibility is not necessarily
the module whose repair yields the greatest benefit
\citep{yoon2026diagnosis}. This distinction motivates our work: rather than
equating responsibility with improvement potential, we evaluate workflow-level
agent priorities through two downstream intervention families.

\paragraph{Optimizing prompts and workflows.}
ProTeGi and TextGrad use textual feedback to improve prompts or compound
computations \citep{pryzant2023protegi,yuksekgonul2025textgrad}.
AFlow searches executable agentic workflows \citep{zhang2025aflow}. HiveMind
uses coalition-based credit for online prompt optimization, while Agents that
Matter studies removal- and model-replacement-based attribution
\citep{xia2026hivemind,lu2026agentsmatter}. Their coalition/removal variants
require a semantics-preserving executor for workflows with absent agents,
which is not consistently available for our workflows; Appendix~\ref{app:baselines}
gives details. Optimas instead optimizes components with globally aligned local
rewards \citep{wu2026optimas}. Our work focuses on the preceding allocation
decision of which agent to improve.

\paragraph{Local response estimation.}
Our response-estimation procedure is methodologically related to local
sensitivity analysis and finite-difference estimation
\citep{borgonovo2016sensitivity,nocedal2006numerical}, as well as local
surrogate methods such as LIME \citep{ribeiro2016lime}. These approaches
characterize local input--output behavior through derivatives, perturbations,
or locally fitted surrogate models. \method{} applies this general principle
at the level of semantically structured agent responses: each perturbed
response is propagated through the downstream workflow, and its effect on
task-level system utility is measured to estimate an agent-level response
signal. This signal is then used to prioritize candidate agents for
improvement under a fixed intervention budget. 
We demonstrate that the estimated priorities translate into practical gains through downstream interventions including prompt optimization and model upgrades.

\section{Conclusion}
We introduced \method{}, a black-box method for deciding where limited
improvement effort should be directed in a fixed LLM-based multi-agent system.
By measuring end-to-end responses to semantic prompt perturbations and fitting
a regularized local response model, \method{} estimates model-implied local
intervention opportunities and identifies agents with promising system-level
improvement potential. Across six workflows, its priorities guide
both prompt optimization and model upgrades, yielding larger selected gains
than uniform agent selection while reducing the expensive downstream calls
required by full-agent interventions. These results suggest that local response
estimation can provide a practical link between failure analysis and
resource-aware multi-agent system improvement. Realized gains depend on the
embedding model, local linearity, probe coverage, and the alignment between
available interventions and the fitted loss-reducing direction. Future work
should translate the continuous response direction into discrete improvements
through direction-conditioned generation, iterative edit-and-measure
procedures, adaptive probing, and uncertainty-aware selection.\label{maintext:end}

\clearpage
\section*{AI use statement}
Generative AI assistants were used in preparing this draft, including literature
retrieval, drafting, code assistance, table construction, and automated
consistency checks. The experiment itself also uses LLMs for workflow
execution, prompt generation, and diagnostic judgments, as distinguished in
the experimental protocol. No missing measurements were synthesized as
results. Responsibility for verification of the experimental evidence,
references, and final submission remains with the authors.

\section*{Ethics statement}
The study uses existing public benchmarks and model services. Generated code
can be unsafe and must be executed in an appropriately restricted environment.
Execution transcripts may contain sensitive material in deployment settings;
only authorized data should be sent to external prompt-generation or judgment
services. Scientific integrity requires retaining infrastructure failures,
distinguishing missing from zero results, and disclosing adaptations of
published baselines.

\section*{Reproducibility statement}
Appendices~\ref{app:setup}--\ref{app:provenance} specify the data roles,
estimator, intervention protocol, statistical units, and artifact details.
Tables are generated from the recorded experimental results and checked for
consistency with the protocol described in the appendix.

\bibliography{references}
\bibliographystyle{iclr2027_conference}

\clearpage
\appendix
\section{Experimental protocol and details}
\label{app:setup}

This appendix specifies the experimental protocol used for all reported
results, including data roles, model configurations, repetitions, statistical
aggregation, baseline adaptations, and cost accounting.

\subsection{Data roles and workflow nodes}
The probing set supplies examples for constructing prompt perturbations,
measuring their effects, and proposing subsequent prompt revisions. The
optimization set scores candidate revisions and selects the candidate to
consider for acceptance. The validation set supplies the additional acceptance
signal specified in Appendix~\ref{app:optimizer}. The test set evaluates the
final intervention and is excluded from agent prioritization and candidate
generation. Following the benchmark partitions used by AFlow, the
probing, optimization, and validation sets originate from separate portions
of each benchmark's development partition. Because the optimization and
validation sets are used adaptively, both are treated as development data.
Model-upgrade experiments
reuse the priorities from the probing set and evaluate gains on the test set,
without using the optimization or validation sets.

\begin{table}[h]
\centering\small
\caption{Sizes of the probing, optimization, validation, and test sets.
The optimization and validation sets are capped within their separate thirds
for ordinary workflows. MATH uses the reduced execution sizes for the
performance experiments; the full-sample row is used only for the cost
scenario in Appendix~\ref{app:cost}.}
\label{tab:partitions}
\begin{tabular}{lrrrrr}
\toprule
Workflow & Probing & Optimization & Validation & Test & Nodes\\
\midrule
HumanEval &45&43&43&33&3\\
MBPP &115&113&113&86&4\\
NQ &268&266&266&266&3\\
DROP &268&266&266&200&2\\
HotpotQA &268&266&266&200&3\\
MATH (reduced execution) &80&80&80&121&5\\
MATH (full-sample cost scenario) &162&161&161&121&5\\
\bottomrule
\end{tabular}
\end{table}

For a development partition of size $N$, the ordinary split starts with
$q=\lfloor N/3\rfloor$. The probing set receives $q+(N-3q)$ examples;
the optimization and validation sets
occupy the second and third blocks of $q$, respectively.
Unused examples above either cap are not reassigned to the other block.
Partition sampling uses seed 42; this is distinct from whether an LLM request
specifies an inference seed. The MATH reduction uses a fixed
80-example subset of the probing set. The cost scenario in Appendix~\ref{app:cost}
instead uses all 162/161/161 probing/optimization/validation examples,
respectively, and all 121 test examples.

\begin{table}[h]
\centering\small
\caption{Node inventory for the six task-specific AFlow-derived workflows. A
named node is the unit of intervention and can execute repeatedly in a trace.}
\begin{tabular}{lp{0.72\linewidth}}
\toprule
Workflow & Intervenable nodes\\
\midrule
HumanEval & CustomCodeGenerate, ScEnsemble, Test\\
MBPP & CodeGenerate, ScEnsemble, Test, FixCode\\
NQ & AnswerGenerate, ScEnsemble, FormatAnswer\\
DROP & Solve, Format\\
HotpotQA & AnswerGenerate, ScEnsemble, FormatAnswer\\
MATH & Programmer, RefineAnswer, DetailedSolution, GenerateSolution, ScEnsemble\\
\bottomrule
\end{tabular}
\end{table}

HumanEval and MBPP judge code by the workflow's execution/testing protocol
\citep{chen2021humaneval,austin2021mbpp}. Our NQ adaptation targets short factual answers
\citep{kwiatkowski2019nq}; HotpotQA involves multi-hop questions
\citep{yang2018hotpotqa}; DROP requires discrete reasoning over passages
\citep{dua2019drop}; MATH assesses competition-style mathematical solutions
\citep{hendrycks2021math}. Each workflow uses the benchmark subset and
workflow-specific scoring implementation described in this appendix. Exact
example IDs, normalizers, the execution harness, and prompt snapshots form
part of the complete replication record.

\subsection{Models, contexts, and repetitions}
The scoring encoder is Qwen3-Embedding-8B with 512 output dimensions
\citep{zhang2025qwenembedding}. The ridge coefficient in
Equation~\ref{eq:ridge} is fixed to $\lambda=1.0$ for every workflow and run.
The local execution model is Qwen3-8B,
with thinking disabled in the intended local request path. The
ordinary configuration uses an 8,192-token context, an input budget of 6,144,
and an output limit of 2,048. MATH uses a 16,384-token context with input
and output budgets of 8,192 each. These are limits rather than measured
response lengths. Only nodes designed to sample multiple candidate answers
use a nonzero execution temperature (0.3 in the prompt-optimization configuration);
other execution nodes use 0. Prompt rewriting uses DeepSeek-V3.2 at 0.7.
Generation and workflow evaluation are different model roles.

Each prioritization algorithm is run independently three times on every
workflow. Prompt-optimization and model-upgrade interventions are each run
independently five times for every agent and dataset; the per-agent gains use
the mean and standard error across these five runs. Method-level selected-gain
and ranking summaries are aggregated over the three prioritization runs. The
five model-upgrade target identities are Qwen3-32B,
Qwen3-235B-A22B-Instruct-2507, Qwen3.7-Plus, DeepSeek-V3.2, and GPT-5 nano.
Qwen3.7-Plus denotes the
service alias recorded during the experiments, with provider and version
metadata retained for reproducibility.
The two additional targets use the OpenRouter identifiers
\texttt{deepseek/deepseek-v3.2} and \texttt{openai/gpt-5-nano}, respectively,
with five test executions per node on each workflow. DeepSeek upgrades disable
reasoning and use the workflow's node-specific temperatures. GPT-5 nano uses low
reasoning effort and a 16,384-token output cap, including reasoning tokens;
its requests omit temperature, top-$p$, and seed.

\section{The ten-slot perturbation specification}
\label{app:perturbations}

Table~\ref{tab:mechanisms} describes the intended mechanism of each formal
slot. Generated prompts are additionally assessed through prompt diffs and
structural guards because a revision may combine multiple semantic changes.
The numerical potential scores used in this paper are \gnorm{}, \iabs{}, and
\meanabs{}; the mechanism names identify intervention types.

\begin{table}[h]
\centering\small
\caption{Formal mechanisms, grouped by construction family. Auxiliary response
measurements support the adaptive construction process.}
\label{tab:mechanisms}
\begin{tabular}{lp{0.61\linewidth}}
\toprule
Mechanism & Intended revision\\
\midrule
Downstream interface & Clarify what the next operator needs and how this
operator must provide it, without changing the workflow's interface.\\
Decomposition control & Make the allocation or ordering of reasoning steps
explicit, keeping the original task and output requirements.\\
Ambiguity resolution & Resolve an underspecified term, priority, or instruction
that could admit incompatible interpretations.\\
Semantic precision & Sharpen the meaning of an existing instruction rather
than merely adding emphasis.\\
Verification & Add or refine a task-relevant internal check while preserving
the required externally visible answer.\\
Edge cases & Clarify boundary conditions or exceptional cases in a
task-general way, not by copying development answers.\\
\midrule
Adaptive counterfactual repair & Use a prior's changed outcomes on the probing set to
identify and address a plausible repair opportunity.\\
Adaptive regression repair & Protect original behavior damaged by an
otherwise useful prior, using development evidence.\\
Adaptive score-conditioned & Revise a prior in light of its measured response
and the development evidence available for that node.\\
\midrule
Sectionwise rephrase & Rephrase editable logical sections separately, keeping
protected fragments and template placeholders intact.\\
\bottomrule
\end{tabular}
\end{table}

Within adaptive construction, up to 36 auxiliary candidates are generated
from 12 initial mechanisms, three slots per mechanism. A single development
evaluation estimates each candidate's loss change on a shared subset of the probing set.
The subset rule uses approximately a quarter of the probing set, at least 40
examples where available, combining low-scoring and random examples without
overlap. MATH's restarted adaptive construction instead uses 32 examples.
These internal measurements are performed once per candidate, not three
times; formal response estimation is a separate repeated step.

Adaptive construction considers up to the five highest-ranked priors with
nonnegative point estimates. A generated adaptive prompt must differ from its
selected prior; repeated duplication advances to the next eligible prior. If
no distinct candidate is produced, the slot is recorded as redundant.
Ordinary mechanisms, sectionwise rephrasing, and adaptive candidates enter the
formal response pool regardless of the sign of their measured effect, and no
test-set screening is used. The resulting pool contains up to ten distinct
prompts per node.

Structural checks preserve placeholders and protected content, require a
changed English prompt with parseable generated fields, and reject unusable
responses. Formal-pool generation and downstream optimizer validation are
different operations with different length policies; the optimizer
policy is specified below. Missing or failed measurements are recorded as
unavailable rather than assigned an observed effect.

\section{Common prompt-optimization intervention}
\label{app:optimizer}

To evaluate all selectors with a common downstream intervention, we use a
shared optimizer inspired by textual-feedback search. The procedure is as
follows.
\begin{enumerate}[leftmargin=*,itemsep=3pt]
\item Hold the original node prompt and its agent-prioritization evidence fixed. Rank eligible
directions by absolute mean effect on the probing set, using lower uncertainty to break
ties. The optimizer uses $|\overline\Delta|\geq0.005$ as its
eligibility threshold. The evidence pool remains unchanged during optimization,
and each direction is considered at most once.
\item For the next direction, provide the original prompt, current accepted
prompt, measured perturbation, a concise LLM-generated description of their
difference, and up to three direction-consistent reversal examples from the probing set.
Positive evidence uses improved examples; negative evidence uses degraded
examples. When none are available, the request omits examples. Expected
answers refer to final task outcomes.
\item Ask DeepSeek-V3.2 for four complete local revisions. All four receive
the same evidence, with different implementations of the relevant change.
Structural validation checks placeholders and protected fragments, rejects
the unchanged current prompt, and applies the recorded length interval from
$0.2$ times the original token length to
$\max(512,3\times\text{original token length})$. Generation is retried up to
three times to obtain four valid candidates; if this is unsuccessful, the
current state is retained and the procedure advances to the next direction.
\item Evaluate each candidate on the optimization set with three repeat-specific evaluations.
Select the lowest mean loss.
Only that candidate can proceed to validation, where the rule below determines
acceptance.
\item On acceptance, update the current state and its optimization- and
validation-set reference losses.
On rejection, retain the previous state. Stop after six accepted updates
or exhaustion of eligible directions. Rejected directions do not count toward
the six-update limit. MATH's reduced record uses three candidates and
at most five attempted directions instead.
\item Hold the final accepted prompt fixed for test-set evaluation. If it
equals the original prompt, use the same per-repeat baseline losses, yielding
zero gain by construction.
\end{enumerate}

\subsection{Acceptance rule}
Let $L_{\mathrm{opt}}^\mathrm{cur}$ be the mean loss on the optimization set
of the current accepted state and $L_{\mathrm{opt}}^*$ that of the beam winner.
For the validation set, let $L_{\mathrm{val}}^\mathrm{orig}$,
$L_{\mathrm{val}}^\mathrm{prev}$, and $L_{\mathrm{val}}^*$ be the mean losses
of the original prompt, previous accepted state, and beam winner, respectively.
Define the optimization gain $o$ and validation gains $v_0$, $v_\mathrm{prev}$,
and $v$ by
\begin{align}
 o &= L_{\mathrm{opt}}^\mathrm{cur}-L_{\mathrm{opt}}^*,\\
 v_0 &= L_{\mathrm{val}}^\mathrm{orig}-L_{\mathrm{val}}^*,\qquad
 v_\mathrm{prev}=L_{\mathrm{val}}^\mathrm{prev}-L_{\mathrm{val}}^*,\\
 v &=\begin{cases}
 v_0,&\text{before the first accepted update},\\
 \min(v_0,v_\mathrm{prev}),&\text{otherwise}.
 \end{cases}
\end{align}
The validation set is evaluated only when $o>-0.005$ (up to numerical tolerance).
The recorded acceptance rule is
\begin{equation}
 \operatorname{accept}(o,v)=
 \begin{cases}
 \mathrm{true},& o>0\ \land\ [v\geq0\ \lor\ o-\max(0,-v)>0.005],\\
 \mathrm{true},& -0.005<o\leq0\ \land\ v>0\ \land\ (o+v)/2>0.0025,\\
 \mathrm{false},&\text{otherwise}.
 \end{cases}
 \label{eq:acceptance}
\end{equation}
The numerical implementation uses an epsilon of $10^{-12}$ at comparisons.
Thresholds are fractions of score, so $0.005$ is 0.5 pp. The second branch
allows improvement on the validation set to compensate for a small decline
on the optimization set.
The final exported prompt is the final accepted state, which need not be
the state with the best optimization-set score alone. The minimum comparison
on the validation set guards
against both the frozen original and the previous accepted state; a first
candidate is not accepted unconditionally. The rule is used for development-set
candidate selection; generalization is evaluated after the final state is
fixed for test-set evaluation.

\subsection{Representative generation instruction}
The following condensed template summarizes the generator's semantics. The
replication artifact will provide the exact serialized prompts and API
parameters used in the experiments.
\begin{quote}\small
You are optimizing one operator's system prompt in a fixed multi-agent
workflow. Lower loss is better. You receive the frozen original prompt,
current accepted prompt, measured perturbation prompt, its signed effect on
the probing set, and a concise description of the change. Any supplied
examples are observations from the probing set; expected answers refer to
the final task outcome.

For positive evidence, incorporate the relevant beneficial difference into
the current prompt. For negative evidence, do not copy the harmful difference;
protect the original behavior it damaged. Add a local safeguard only when
the evidence supports a task-general risk. Do not invent an opposite rule
merely because the measured direction is negative.

Return four complete, distinct local revisions using the same full evidence.
Preserve every required placeholder and protected fragment. Do not summarize
or globally rewrite the current prompt. Do not copy example-specific answers.
Vary how the supported change is implemented, not which evidence is visible.
Output only a JSON list of four prompt strings, without commentary or fences.
\end{quote}

\section{Statistical units and recomputation}
\label{app:statistics}

For prioritization run $r\in\{1,2,3\}$, let
$A_r=\arg\max_i S_i^{(r)}$ be the set of top-tied nodes. The reported selected
gain for that group is
$g_r=100|A_r|^{-1}\sum_{i\in A_r}\widehat U_i$.
This quantity is the expected gain under uniform tie-breaking. Kendall $t_r$ compares
the complete score vector with the vector of mean intervention gains.
For either $z=g$ or $z=t$, the three-run summary is
\begin{equation}
 \overline z=\frac{1}{3}\sum_{r=1}^{3}z_r,\qquad
 \SEsel(z)=\frac{1}{\sqrt{3}}
 \sqrt{\frac{\sum_r(z_r-\overline z)^2}{2}}.
\end{equation}
This aggregation describes variability across the three independent
prioritization runs.

We report Kendall against the vector of mean intervention gains. For each
node and intervention, the five repeated gains $u_1,\ldots,u_5$ determine the
mean intervention gain and its standard error. Sample variance uses divisor
$T-1$ with $T=5$, and $\SEtest=\sqrt{s^2/T}$.

The uniform-agent reference first averages nodes within each repeat and
then estimates its SE across repeats; individual node SEs cannot be averaged
to reconstruct it because covariance matters. In the main model-upgrade table,
the Qwen panel averages the selected-gain and Kendall metrics across the three
Qwen targets. For the ablations, we instead average all five target models'
per-node gains \emph{before} computing selected gain and Kendall.

The six-workflow macro statistics weight workflows equally, not individual
examples or nodes. A two-node workflow such as DROP has only one pair;
perfect Kendall on it is less demanding than a complete five-node ordering.
Point estimates are interpreted in light of sample sizes, shared baselines,
and ties; standard-error bars summarize variability rather than define a
standalone significance test.

\section{All-node gain references and complete uncertainty tables}
\label{app:nodegains}
\begin{table}[ht]
\caption{Per-node mean test gains (pp) for prompt optimization and all five
model-upgrade targets used as evaluation references,
rounded to three decimals. The listed nodes cover all six workflows. MeanGain
averages each node's five
model-upgrade means before ranking. Each prompt-optimization and model-upgrade
entry averages five intervention runs.
No uncertainty is implied by this mean-only table.}
\label{tab:all-agent-gains}
\centering
\footnotesize
\setlength{\tabcolsep}{3pt}
\resizebox{\textwidth}{!}{%
\begin{tabular}{llrrrrrrr}
\toprule
& & & \multicolumn{6}{c}{Model-upgrade target / average} \\
\cmidrule(lr){4-9}
Workflow & Node & \shortstack{Prompt\\optimization} & 32B & 3.7-plus & 235B & \shortstack{DeepSeek\\V3.2} & \shortstack{GPT-5\\nano} & MeanGain \\
\midrule
HumanEval & \textsf{CustomCodeGenerate} & $+4.242$ & $+15.152$ & $+27.273$ & $+24.242$ & $+18.788$ & $+24.849$ & $+22.061$ \\
 & \textsf{ScEnsemble} & $+1.212$ & $+0.606$ & $+0.000$ & $+0.000$ & $+0.000$ & $+0.606$ & $+0.242$ \\
 & \textsf{Test} & $+4.849$ & $+4.242$ & $+9.091$ & $+9.091$ & $+9.697$ & $+8.485$ & $+8.121$ \\
\addlinespace[3pt]
MBPP & \textsf{CodeGenerate} & $+3.023$ & $+6.744$ & $+9.070$ & $+7.209$ & $+8.605$ & $+2.093$ & $+6.744$ \\
 & \textsf{ScEnsemble} & $+0.000$ & $-0.233$ & $-0.233$ & $-0.465$ & $+0.233$ & $-0.465$ & $-0.233$ \\
 & \textsf{Test} & $+7.907$ & $+7.442$ & $+21.168$ & $+11.628$ & $+12.325$ & $+17.210$ & $+13.955$ \\
 & \textsf{FixCode} & $+0.465$ & $+0.930$ & $+8.553$ & $+3.488$ & $+2.325$ & $+3.954$ & $+3.850$ \\
\addlinespace[3pt]
NQ & \textsf{AnswerGenerate} & $+2.141$ & $+10.634$ & $+19.780$ & $+21.782$ & $+23.085$ & $+14.097$ & $+17.876$ \\
 & \textsf{ScEnsemble} & $-0.111$ & $-0.305$ & $-0.122$ & $-0.984$ & $+1.705$ & $+2.138$ & $+0.486$ \\
 & \textsf{FormatAnswer} & $+2.002$ & $+1.298$ & $+1.326$ & $+2.630$ & $+1.881$ & $+1.343$ & $+1.696$ \\
\addlinespace[3pt]
DROP & \textsf{Solve} & $+4.994$ & $+6.633$ & $+12.822$ & $+11.126$ & $+8.969$ & $+6.503$ & $+9.211$ \\
 & \textsf{Format} & $-1.038$ & $+3.425$ & $+7.448$ & $+7.346$ & $+4.947$ & $+2.658$ & $+5.165$ \\
\addlinespace[3pt]
HotpotQA & \textsf{AnswerGenerate} & $+1.534$ & $+5.739$ & $+10.463$ & $+7.878$ & $+5.182$ & $+3.033$ & $+6.459$ \\
 & \textsf{ScEnsemble} & $-0.547$ & $+0.403$ & $+0.145$ & $+0.533$ & $+1.472$ & $+1.378$ & $+0.786$ \\
 & \textsf{FormatAnswer} & $+0.000$ & $+0.359$ & $+2.314$ & $+2.248$ & $+2.387$ & $+0.116$ & $+1.485$ \\
\addlinespace[3pt]
MATH & \textsf{Programmer} & $+1.377$ & $+1.653$ & $+5.234$ & $+4.408$ & $+2.204$ & $+1.378$ & $+2.975$ \\
 & \textsf{RefineAnswer} & $+1.551$ & $+1.377$ & $+11.771$ & $+1.653$ & $+4.187$ & $+7.493$ & $+5.296$ \\
 & \textsf{DetailedSolution} & $+1.728$ & $+0.551$ & $+8.264$ & $+9.091$ & $+3.526$ & $+6.832$ & $+5.653$ \\
 & \textsf{GenerateSolution} & $+2.002$ & $+4.959$ & $+13.751$ & $+13.499$ & $+12.286$ & $+13.113$ & $+11.522$ \\
 & \textsf{ScEnsemble} & $-0.755$ & $-1.423$ & $+3.581$ & $+3.581$ & $+4.187$ & $+0.551$ & $+2.096$ \\
\bottomrule
\end{tabular}%
}
\end{table}

The all-node outcomes provide offline reference values for evaluating agent
selection. They are measured intervention gains and are excluded from all
selector inputs.

\begin{table}[t]
\caption{Expanded view of the \method{} uncertainty in Table~\ref{tab:rq3}.
Every $\pm$ is \SEsel{} across the three prioritization runs against the
per-node mean gains computed from five intervention runs. It is not
\SEtest{}. Zero \SEsel{} can
result from identical selections despite nonzero test execution variability.
The table covers all five upgrade targets.
Target aliases match Table~\ref{tab:rq3}.}
\label{tab:rq3-uncertainty}
\centering
\scriptsize
\setlength{\tabcolsep}{2.5pt}
\begin{tabular}{llcc@{\hspace{8pt}}llcc}
\toprule
Workflow & Target & \gnorm{} gain (pp) & \gnorm{} $\tau_b$
& Workflow & Target & \gnorm{} gain (pp) & \gnorm{} $\tau_b$ \\
\midrule
HumanEval & 32B & \gainse{+11.515}{3.636} & \gainse{0.778}{0.222}
& DROP & 32B & \gainse{+6.633}{0.000} & \gainse{1.000}{0.000} \\
 & 3.7-plus & \gainse{+21.212}{6.061} & \gainse{0.778}{0.222}
& & 3.7-plus & \gainse{+12.822}{0.000} & \gainse{1.000}{0.000} \\
 & 235B & \gainse{+19.192}{5.051} & \gainse{0.778}{0.222}
& & 235B & \gainse{+11.126}{0.000} & \gainse{1.000}{0.000} \\
 & DeepSeek-V3.2 & \gainse{+15.758}{3.030} & \gainse{0.778}{0.222}
& & DeepSeek-V3.2 & \gainse{+8.969}{0.000} & \gainse{1.000}{0.000} \\
 & GPT-5 nano & \gainse{+19.394}{5.455} & \gainse{0.778}{0.222}
& & GPT-5 nano & \gainse{+6.503}{0.000} & \gainse{1.000}{0.000} \\
\addlinespace[3pt]
MBPP & 32B & \gainse{+7.442}{0.000} & \gainse{0.889}{0.111}
& HotpotQA & 32B & \gainse{+5.739}{0.000} & \gainse{0.778}{0.222} \\
 & 3.7-plus & \gainse{+21.168}{0.000} & \gainse{0.889}{0.111}
& & 3.7-plus & \gainse{+10.463}{0.000} & \gainse{0.556}{0.222} \\
 & 235B & \gainse{+11.628}{0.000} & \gainse{0.889}{0.111}
& & 235B & \gainse{+7.878}{0.000} & \gainse{0.556}{0.222} \\
 & DeepSeek-V3.2 & \gainse{+12.325}{0.000} & \gainse{0.889}{0.111}
& & DeepSeek-V3.2 & \gainse{+5.182}{0.000} & \gainse{0.556}{0.222} \\
 & GPT-5 nano & \gainse{+17.210}{0.000} & \gainse{0.556}{0.111}
& & GPT-5 nano & \gainse{+3.033}{0.000} & \gainse{0.778}{0.222} \\
\addlinespace[3pt]
NQ & 32B & \gainse{+10.634}{0.000} & \gainse{0.778}{0.222}
& MATH & 32B & \gainse{+3.765}{1.194} & \gainse{0.600}{0.115} \\
 & 3.7-plus & \gainse{+19.780}{0.000} & \gainse{0.778}{0.222}
& & 3.7-plus & \gainse{+13.091}{0.660} & \gainse{0.733}{0.176} \\
 & 235B & \gainse{+21.782}{0.000} & \gainse{0.778}{0.222}
& & 235B & \gainse{+9.550}{3.949} & \gainse{0.400}{0.115} \\
 & DeepSeek-V3.2 & \gainse{+23.085}{0.000} & \gainse{0.778}{0.222}
& & DeepSeek-V3.2 & \gainse{+9.587}{2.700} & \gainse{0.246}{0.186} \\
 & GPT-5 nano & \gainse{+14.097}{0.000} & \gainse{0.556}{0.222}
& & GPT-5 nano & \gainse{+11.240}{1.873} & \gainse{0.733}{0.176} \\
\bottomrule
\end{tabular}
\end{table}

\begin{table}[p]
\caption{Agent rankings by mean model-upgrade test gain for HumanEval, MBPP, NQ.
Each row ranks nodes from highest to lowest gain for one target model.
Parentheses give mean gains in percentage points; $=$ denotes a tie.
Gains use the references in Table~\ref{tab:all-agent-gains}; rankings are computed before display rounding.}
\label{tab:agent-gain-rankings-1}
\centering
\scriptsize
\setlength{\tabcolsep}{3pt}
\renewcommand{\arraystretch}{1.00}
\begin{tabular}{@{}>{\raggedright\arraybackslash}p{0.12\textwidth}>{\raggedright\arraybackslash}p{0.23\textwidth}>{\raggedright\arraybackslash}p{0.60\textwidth}@{}}
\toprule
Workflow & Upgrade target & Agent ranking (mean gain, pp) \\
\midrule
HumanEval & Qwen3-32B & \textsf{CustomCodeGenerate} (+15.152) $>$ \textsf{Test} (+4.242) \newline $>$ \textsf{ScEnsemble} (+0.606) \\
 & Qwen3-235B-A22B & \textsf{CustomCodeGenerate} (+24.242) $>$ \textsf{Test} (+9.091) \newline $>$ \textsf{ScEnsemble} (+0.000) \\
 & Qwen3.7-Plus & \textsf{CustomCodeGenerate} (+27.273) $>$ \textsf{Test} (+9.091) \newline $>$ \textsf{ScEnsemble} (+0.000) \\
 & DeepSeek-V3.2 & \textsf{CustomCodeGenerate} (+18.788) $>$ \textsf{Test} (+9.697) \newline $>$ \textsf{ScEnsemble} (+0.000) \\
 & GPT-5 nano & \textsf{CustomCodeGenerate} (+24.849) $>$ \textsf{Test} (+8.485) \newline $>$ \textsf{ScEnsemble} (+0.606) \\
\addlinespace[5pt]
MBPP & Qwen3-32B & \textsf{Test} (+7.442) $>$ \textsf{CodeGenerate} (+6.744) \newline $>$ \textsf{FixCode} (+0.930) $>$ \textsf{ScEnsemble} (-0.233) \\
 & Qwen3-235B-A22B & \textsf{Test} (+11.628) $>$ \textsf{CodeGenerate} (+7.209) \newline $>$ \textsf{FixCode} (+3.488) $>$ \textsf{ScEnsemble} (-0.465) \\
 & Qwen3.7-Plus & \textsf{Test} (+21.168) $>$ \textsf{CodeGenerate} (+9.070) \newline $>$ \textsf{FixCode} (+8.553) $>$ \textsf{ScEnsemble} (-0.233) \\
 & DeepSeek-V3.2 & \textsf{Test} (+12.325) $>$ \textsf{CodeGenerate} (+8.605) \newline $>$ \textsf{FixCode} (+2.325) $>$ \textsf{ScEnsemble} (+0.233) \\
 & GPT-5 nano & \textsf{Test} (+17.210) $>$ \textsf{FixCode} (+3.954) \newline $>$ \textsf{CodeGenerate} (+2.093) $>$ \textsf{ScEnsemble} (-0.465) \\
\addlinespace[5pt]
NQ & Qwen3-32B & \textsf{AnswerGenerate} (+10.634) $>$ \textsf{FormatAnswer} (+1.298) \newline $>$ \textsf{ScEnsemble} (-0.305) \\
 & Qwen3-235B-A22B & \textsf{AnswerGenerate} (+21.782) $>$ \textsf{FormatAnswer} (+2.630) \newline $>$ \textsf{ScEnsemble} (-0.984) \\
 & Qwen3.7-Plus & \textsf{AnswerGenerate} (+19.780) $>$ \textsf{FormatAnswer} (+1.326) \newline $>$ \textsf{ScEnsemble} (-0.122) \\
 & DeepSeek-V3.2 & \textsf{AnswerGenerate} (+23.085) $>$ \textsf{FormatAnswer} (+1.881) \newline $>$ \textsf{ScEnsemble} (+1.705) \\
 & GPT-5 nano & \textsf{AnswerGenerate} (+14.097) $>$ \textsf{ScEnsemble} (+2.138) \newline $>$ \textsf{FormatAnswer} (+1.343) \\
\bottomrule
\end{tabular}
\end{table}

\begin{table}[p]
\caption{Agent rankings by mean model-upgrade test gain for DROP, HotpotQA, MATH.
Each row ranks nodes from highest to lowest gain for one target model.
Parentheses give mean gains in percentage points; $=$ denotes a tie.
Gains use the references in Table~\ref{tab:all-agent-gains}; rankings are computed before display rounding.}
\label{tab:agent-gain-rankings-2}
\centering
\scriptsize
\setlength{\tabcolsep}{3pt}
\renewcommand{\arraystretch}{1.00}
\begin{tabular}{@{}>{\raggedright\arraybackslash}p{0.12\textwidth}>{\raggedright\arraybackslash}p{0.23\textwidth}>{\raggedright\arraybackslash}p{0.60\textwidth}@{}}
\toprule
Workflow & Upgrade target & Agent ranking (mean gain, pp) \\
\midrule
DROP & Qwen3-32B & \textsf{Solve} (+6.633) $>$ \textsf{Format} (+3.425) \\
 & Qwen3-235B-A22B & \textsf{Solve} (+11.126) $>$ \textsf{Format} (+7.346) \\
 & Qwen3.7-Plus & \textsf{Solve} (+12.822) $>$ \textsf{Format} (+7.448) \\
 & DeepSeek-V3.2 & \textsf{Solve} (+8.969) $>$ \textsf{Format} (+4.947) \\
 & GPT-5 nano & \textsf{Solve} (+6.503) $>$ \textsf{Format} (+2.658) \\
\addlinespace[5pt]
HotpotQA & Qwen3-32B & \textsf{AnswerGenerate} (+5.739) $>$ \textsf{ScEnsemble} (+0.403) \newline $>$ \textsf{FormatAnswer} (+0.359) \\
 & Qwen3-235B-A22B & \textsf{AnswerGenerate} (+7.878) $>$ \textsf{FormatAnswer} (+2.248) \newline $>$ \textsf{ScEnsemble} (+0.533) \\
 & Qwen3.7-Plus & \textsf{AnswerGenerate} (+10.463) $>$ \textsf{FormatAnswer} (+2.314) \newline $>$ \textsf{ScEnsemble} (+0.145) \\
 & DeepSeek-V3.2 & \textsf{AnswerGenerate} (+5.182) $>$ \textsf{FormatAnswer} (+2.387) \newline $>$ \textsf{ScEnsemble} (+1.472) \\
 & GPT-5 nano & \textsf{AnswerGenerate} (+3.033) $>$ \textsf{ScEnsemble} (+1.378) \newline $>$ \textsf{FormatAnswer} (+0.116) \\
\addlinespace[5pt]
MATH & Qwen3-32B & \textsf{GenerateSolution} (+4.959) $>$ \textsf{Programmer} (+1.653) \newline $>$ \textsf{RefineAnswer} (+1.377) $>$ \textsf{DetailedSolution} (+0.551) \newline $>$ \textsf{ScEnsemble} (-1.423) \\
 & Qwen3-235B-A22B & \textsf{GenerateSolution} (+13.499) $>$ \textsf{DetailedSolution} (+9.091) \newline $>$ \textsf{Programmer} (+4.408) $>$ \textsf{ScEnsemble} (+3.581) \newline $>$ \textsf{RefineAnswer} (+1.653) \\
 & Qwen3.7-Plus & \textsf{GenerateSolution} (+13.751) $>$ \textsf{RefineAnswer} (+11.771) \newline $>$ \textsf{DetailedSolution} (+8.264) $>$ \textsf{Programmer} (+5.234) \newline $>$ \textsf{ScEnsemble} (+3.581) \\
 & DeepSeek-V3.2 & \textsf{GenerateSolution} (+12.286) $>$ \textsf{RefineAnswer} (+4.187) \newline $=$ \textsf{ScEnsemble} (+4.187) $>$ \textsf{DetailedSolution} (+3.526) \newline $>$ \textsf{Programmer} (+2.204) \\
 & GPT-5 nano & \textsf{GenerateSolution} (+13.113) $>$ \textsf{RefineAnswer} (+7.493) \newline $>$ \textsf{DetailedSolution} (+6.832) $>$ \textsf{Programmer} (+1.378) \newline $>$ \textsf{ScEnsemble} (+0.551) \\
\bottomrule
\end{tabular}
\end{table}

\begingroup
\showresultsetrue

\endgroup
\clearpage




\section{External selectors and adaptations}
\label{app:baselines}

The six external selectors use the same frozen DEV-A trajectories, node
inventory, and downstream gain matrices. Diagnostic judgments use
DeepSeek-V3.2, while task execution uses the base workflow model. Each selector
is run independently three times. Valid task-level attributions are aggregated
over failed development trajectories to form a workflow-level node ranking;
unparseable outputs and explicit abstentions contribute no vote. The resulting
ranking is used unchanged for both prompt optimization and model upgrades.

\paragraph{Who\&When-AAO.}
We use the paper's All-at-Once prompt, which reads the complete failed trajectory
in one request and returns an agent, an error step, and a reason. A valid
agent--step attribution contributes one vote to that agent. Votes are summed
over failed trajectories to obtain the node ranking.

\paragraph{A2P.}
One diagnostic request performs the paper's three reasoning stages: abduction
of the failure cause, a minimal corrective action, and prediction of the
resulting continuation. The final agent--step attribution contributes one vote,
and the workflow-level ranking is obtained by aggregating these votes.

\paragraph{Adapted locator.}
We first apply the task critic adapted from the paper's critic template to the
final workflow output and its recorded evaluation evidence. The locator then
uses this feedback together with agent roles and the complete trajectory to
return a set of weak agents. Each trajectory contributes one vote divided
uniformly over that set, and the accumulated shares determine the ranking. We
evaluate this localization component independently of the paper's subsequent
workflow-optimization stage.

\paragraph{AgentRx.}
The recorded trajectory is converted to AgentRx's canonical intermediate
representation. Its static and dynamic invariant generators produce checks,
the generated Python and natural-language checks are evaluated, and the native
judge assigns a failure category and responsible step from the resulting
violation log. Generated Python checks run in isolated containers. A mapped
responsible step contributes one vote to its owning agent; intermediate
violation counts are not used as node scores.

\paragraph{CHIEF.}
The implementation follows CHIEF's staged diagnosis: retrieval-augmented task
decomposition, OTAR construction, subtask/agent/step dependency graphs, virtual
oracle construction, hierarchical backtracking, and the final
Local--Planning-Control--Data-Flow--Deviation-Aware attribution. The final
supported agent receives one task-level vote. The RAG store and the complete
intermediate diagnosis are retained for every trajectory.

\paragraph{FAMAS.}
For each failed anchor trajectory, the reported configuration combines the
anchor with 20 additional replays and retains at most ten native leaf events
per trajectory. The implementation
abstracts events into agent/action/state triples, clusters semantically
equivalent action--state pairs within each agent, and applies the FAMAS
frequency, coverage, and success/failure spectrum score with decay
$\lambda=0.9$. The owner of the unique highest-scoring anchor triple receives
one vote; a top-score tie produces an abstention.

\paragraph{Why coalition/removal baselines are not included.}
HiveMind assigns node credit from coalition values on a workflow DAG using a
Shapley calculation \citep{xia2026hivemind}. Agents that Matter likewise
formalizes agent credit through explicit removal protocols, alongside a
separate model-replacement analysis \citep{lu2026agentsmatter}. In the five chain-structured
workflows used here, the current executable interface admits only the complete
source-to-sink coalition: removing an intermediate node leaves a downstream
node without its required input. Under this coalition set, every node receives
the same value $v(N)/|N|$, so the selected node is determined only by tie
breaking. MATH admits additional graph-valid coalitions, but its composite
operators do not define what a downstream node should receive when one of its
required parents is absent. Filling those inputs with null values or replaying
outputs from the complete workflow would define a different intervention rather
than execute the coalition being evaluated. We therefore do not include the
coalition/removal variants of these methods in the matched selector comparison,
because the six workflows do not provide a common semantics-preserving removal
operator.

Tables~\ref{tab:rq2} and~\ref{tab:rq3} report the comparison under the shared
aggregation and intervention protocol above.

\section{Scoring and perturbation-budget ablations}
\label{app:budget}
Using the signed probe effect
$\Delta_{im}=\overline L_0-\overline L_i^{(m)}$, the two magnitude-only
scoring baselines are
\begin{equation}
 S_i^{\mathrm{abs}}=\max_m |\Delta_{im}|,
 \qquad
 S_i^{\mathrm{mean}}=\frac{1}{M_i}\sum_m|\Delta_{im}|.
 \label{eq:ablations}
\end{equation}
These scores ignore whether a perturbation helps or harms and do not use the
embedding direction or displacement size.

Table~\ref{tab:abl-macro} reports completed variants with three, five, and up
to ten formal probes per node. All variants use the same datasets, execution
models, downstream interventions, and evaluation rules. The nominal probe
budget does not include auxiliary evaluations used during adaptive
construction, which are retained in the reduced-probe variants.

Tables~\ref{tab:abl-score-detail} and~\ref{tab:abl-budget-detail} report the
corresponding results for each workflow. Gains are in percentage points;
each $\pm$ term is the standard error across the three prioritization runs.
The model-upgrade columns first average each node's gain over the five target
models. The macro rows use equal dataset weights and reproduce
Table~\ref{tab:abl-macro}.

\begin{table}[!htbp]
\caption{Per-workflow scoring-rule ablation. Each cell reports the mean
$\pm\SEsel{}$ over three prioritization runs. Model-upgrade gains first average
each node over all five target models.}
\label{tab:abl-score-detail}
\centering
\scriptsize
\setlength{\tabcolsep}{3.2pt}
\begin{tabular}{llcccc}
\toprule
& & \multicolumn{2}{c}{Prompt optimization}
& \multicolumn{2}{c}{Model upgrades} \\
\cmidrule(lr){3-4}\cmidrule(lr){5-6}
Workflow & Score & Gain (pp) $\uparrow$ & $\tau_b\uparrow$
& Gain (pp) $\uparrow$ & $\tau_b\uparrow$ \\
\midrule
HumanEval & \gnorm{}  & \gainse{+4.444}{0.202} & \gainse{0.556}{0.222} & \gainse{+17.414}{4.647} & \gainse{0.778}{0.222} \\
 & \iabs{}            & \gainse{+4.849}{0.000} & \gainse{1.000}{0.000} & \gainse{+8.121}{0.000}  & \gainse{0.333}{0.000} \\
 & \meanabs{}         & \gainse{+4.849}{0.000} & \gainse{1.000}{0.000} & \gainse{+8.121}{0.000}  & \gainse{0.333}{0.000} \\
\addlinespace[2pt]
MBPP & \gnorm{}       & \gainse{+7.907}{0.000} & \gainse{0.889}{0.111} & \gainse{+13.955}{0.000} & \gainse{0.889}{0.111} \\
 & \iabs{}            & \gainse{+4.651}{1.628} & \gainse{0.778}{0.111} & \gainse{+9.148}{2.404}  & \gainse{0.778}{0.111} \\
 & \meanabs{}         & \gainse{+6.279}{1.628} & \gainse{0.556}{0.111} & \gainse{+11.551}{2.404} & \gainse{0.556}{0.111} \\
\addlinespace[2pt]
NQ & \gnorm{}         & \gainse{+2.141}{0.000} & \gainse{0.778}{0.222} & \gainse{+17.876}{0.000} & \gainse{0.778}{0.222} \\
 & \iabs{}            & \gainse{+2.141}{0.000} & \gainse{0.778}{0.222} & \gainse{+17.876}{0.000} & \gainse{0.778}{0.222} \\
 & \meanabs{}         & \gainse{+2.095}{0.046} & \gainse{0.778}{0.222} & \gainse{+12.483}{5.393} & \gainse{0.778}{0.222} \\
\addlinespace[2pt]
DROP & \gnorm{}       & \gainse{+4.994}{0.000} & \gainse{1.000}{0.000} & \gainse{+9.211}{0.000} & \gainse{1.000}{0.000} \\
 & \iabs{}            & \gainse{+4.994}{0.000} & \gainse{1.000}{0.000} & \gainse{+9.211}{0.000} & \gainse{1.000}{0.000} \\
 & \meanabs{}         & \gainse{+4.994}{0.000} & \gainse{1.000}{0.000} & \gainse{+9.211}{0.000} & \gainse{1.000}{0.000} \\
\addlinespace[2pt]
HotpotQA & \gnorm{}   & \gainse{+1.534}{0.000} & \gainse{0.556}{0.222} & \gainse{+6.459}{0.000} & \gainse{0.556}{0.222} \\
 & \iabs{}            & \gainse{+1.534}{0.000} & \gainse{0.556}{0.222} & \gainse{+6.459}{0.000} & \gainse{0.556}{0.222} \\
 & \meanabs{}         & \gainse{+1.534}{0.000} & \gainse{0.778}{0.222} & \gainse{+6.459}{0.000} & \gainse{0.778}{0.222} \\
\addlinespace[2pt]
MATH & \gnorm{}       & \gainse{+1.852}{0.150} & \gainse{0.667}{0.067} & \gainse{+9.447}{2.075} & \gainse{0.667}{0.067} \\
 & \iabs{}            & \gainse{+1.485}{0.200} & \gainse{0.399}{0.204} & \gainse{+5.563}{0.052} & \gainse{0.399}{0.204} \\
 & \meanabs{}         & \gainse{+1.702}{0.181} & \gainse{0.533}{0.176} & \gainse{+6.717}{2.524} & \gainse{0.533}{0.176} \\
\midrule
Macro & \gnorm{}      & $+3.812$ & $0.741$ & $+12.394$ & $0.778$ \\
 & \iabs{}            & $+3.276$ & $0.752$ & $+9.396$ & $0.641$ \\
 & \meanabs{}         & $+3.576$ & $0.774$ & $+9.090$ & $0.663$ \\
\bottomrule
\end{tabular}
\end{table}

\begin{table}[!htbp]
\caption{Per-workflow perturbation-budget comparison using \gnorm{}. Each cell
reports the mean $\pm\SEsel{}$ over three prioritization runs. The full setting
uses at most ten formal probes per node.}
\label{tab:abl-budget-detail}
\centering
\scriptsize
\setlength{\tabcolsep}{3.2pt}
\begin{tabular}{llcccc}
\toprule
& & \multicolumn{2}{c}{Prompt optimization}
& \multicolumn{2}{c}{Model upgrades} \\
\cmidrule(lr){3-4}\cmidrule(lr){5-6}
Workflow & Probes & Gain (pp) $\uparrow$ & $\tau_b\uparrow$
& Gain (pp) $\uparrow$ & $\tau_b\uparrow$ \\
\midrule
HumanEval & 3    & \gainse{+4.849}{0.000} & \gainse{1.000}{0.000} & \gainse{+8.121}{0.000}  & \gainse{0.333}{0.000} \\
 & 5             & \gainse{+4.849}{0.000} & \gainse{1.000}{0.000} & \gainse{+8.121}{0.000}  & \gainse{0.333}{0.000} \\
 & Full          & \gainse{+4.444}{0.202} & \gainse{0.556}{0.222} & \gainse{+17.414}{4.647} & \gainse{0.778}{0.222} \\
\addlinespace[2pt]
MBPP & 3         & \gainse{+5.426}{2.481} & \gainse{0.667}{0.192} & \gainse{+10.587}{3.368} & \gainse{0.667}{0.192} \\
 & 5             & \gainse{+5.426}{2.481} & \gainse{0.667}{0.192} & \gainse{+10.587}{3.368} & \gainse{0.667}{0.192} \\
 & Full          & \gainse{+7.907}{0.000} & \gainse{0.889}{0.111} & \gainse{+13.955}{0.000} & \gainse{0.889}{0.111} \\
\addlinespace[2pt]
NQ & 3           & \gainse{+1.344}{0.728} & \gainse{0.111}{0.588} & \gainse{+6.686}{5.606}  & \gainse{0.111}{0.588} \\
 & 5             & \gainse{+2.095}{0.046} & \gainse{0.333}{0.385} & \gainse{+12.483}{5.393} & \gainse{0.333}{0.385} \\
 & Full          & \gainse{+2.141}{0.000} & \gainse{0.778}{0.222} & \gainse{+17.876}{0.000} & \gainse{0.778}{0.222} \\
\addlinespace[2pt]
DROP & 3         & \gainse{+0.973}{2.011} & \gainse{-0.333}{0.667} & \gainse{+6.514}{1.349} & \gainse{-0.333}{0.667} \\
 & 5             & \gainse{+2.984}{2.011} & \gainse{0.333}{0.667}  & \gainse{+7.862}{1.349} & \gainse{0.333}{0.667} \\
 & Full          & \gainse{+4.994}{0.000} & \gainse{1.000}{0.000}  & \gainse{+9.211}{0.000} & \gainse{1.000}{0.000} \\
\addlinespace[2pt]
HotpotQA & 3     & \gainse{+0.511}{0.511} & \gainse{-0.111}{0.222} & \gainse{+3.143}{1.658} & \gainse{-0.111}{0.222} \\
 & 5             & \gainse{+0.511}{0.511} & \gainse{-0.111}{0.222} & \gainse{+3.143}{1.658} & \gainse{-0.111}{0.222} \\
 & Full          & \gainse{+1.534}{0.000} & \gainse{0.556}{0.222}  & \gainse{+6.459}{0.000} & \gainse{0.556}{0.222} \\
\addlinespace[2pt]
MATH & 3         & \gainse{+1.610}{0.059} & \gainse{0.400}{0.115} & \gainse{+5.415}{0.119} & \gainse{0.400}{0.115} \\
 & 5             & \gainse{+1.552}{0.101} & \gainse{0.267}{0.176} & \gainse{+4.641}{0.840} & \gainse{0.267}{0.176} \\
 & Full          & \gainse{+1.852}{0.150} & \gainse{0.667}{0.067} & \gainse{+9.447}{2.075} & \gainse{0.667}{0.067} \\
\midrule
Macro & 3        & $+2.452$ & $0.289$ & $+6.744$ & $0.178$ \\
 & 5             & $+2.903$ & $0.415$ & $+7.806$ & $0.304$ \\
 & Full          & $+3.812$ & $0.741$ & $+12.394$ & $0.778$ \\
\bottomrule
\end{tabular}
\end{table}

\FloatBarrier

\section{Cost accounting}
\label{app:cost}

All tables in this section report measured execution costs over the six
datasets, aggregated from the recorded execution ledgers. Token counts include
input and output tokens and are reported in millions (M); fractional request
counts can result from averaging measured runs and weighted selections. Each
cell shows the absolute
cost above and, in parentheses, the percentage relative to \method{} in the
corresponding column and setting; \method{} is therefore 100\% by definition.

\begin{table}[!htbp]
\centering\footnotesize
\caption{Cost of one agent-prioritization run over all six datasets.}
\label{tab:prioritization-cost}
\setlength{\tabcolsep}{4pt}
\begin{tabular}{lcccc}
\toprule
Method & \shortstack{Qwen3-8B\\requests}
& \shortstack{Qwen tokens\\(M)}
& \shortstack{DeepSeek-V3.2\\requests}
& \shortstack{DeepSeek tokens\\(M)}\\
\midrule
\textbf{IF-MAP} & \paircell{63{,}088}{100.00\%} & \paircell{47.141}{100.00\%}
& \paircell{662}{100.00\%} & \paircell{3.049}{100.00\%}\\
Who\&When-AAO & \paircell{35{,}288}{55.93\%} & \paircell{26.672}{56.58\%}
& \paircell{1{,}943}{293.50\%} & \paircell{18.341}{601.54\%}\\
A2P & \paircell{35{,}288}{55.93\%} & \paircell{26.672}{56.58\%}
& \paircell{1{,}938}{292.75\%} & \paircell{18.645}{611.51\%}\\
Adapted locator & \paircell{35{,}288}{55.93\%} & \paircell{26.672}{56.58\%}
& \paircell{3{,}197}{482.93\%} & \paircell{21.138}{693.28\%}\\
AgentRx & \paircell{35{,}288}{55.93\%} & \paircell{26.672}{56.58\%}
& \paircell{8{,}506}{1{,}284.89\%} & \paircell{202.576}{6{,}644.01\%}\\
CHIEF & \paircell{35{,}288}{55.93\%} & \paircell{26.672}{56.58\%}
& \paircell{10{,}225}{1{,}544.56\%} & \paircell{142.013}{4{,}657.69\%}\\
FAMAS & \paircell{72{,}894}{115.54\%} & \paircell{52.709}{111.81\%}
& \paircell{18{,}228}{2{,}753.47\%} & \paircell{171.566}{5{,}626.96\%}\\
\bottomrule
\end{tabular}
\end{table}

\begin{table}[!htbp]
\centering\footnotesize
\caption{Prompt-optimization cost over all six datasets. \method{} first
selects an agent and optimizes only that agent; Full-Agent PromptOpt applies
the same custom prompt optimizer to every agent.}
\label{tab:prompt-cost}
\setlength{\tabcolsep}{5pt}
\begin{tabular}{lcccc}
\toprule
Method & \shortstack{Qwen3-8B\\requests}
& \shortstack{Qwen tokens\\(M)}
& \shortstack{DeepSeek-V3.2\\(TextGrad)\\requests}
& \shortstack{DeepSeek (TextGrad)\\tokens (M)}\\
\midrule
\textbf{IF-MAP} & \paircell{203{,}546.74}{100.00\%}
& \paircell{352.994}{100.00\%} & \paircell{950.00}{100.00\%}
& \paircell{4.249}{100.00\%}\\
Full-Agent PromptOpt & \paircell{663{,}341.97}{325.89\%}
& \paircell{1{,}366.978}{387.25\%} & \paircell{1{,}662.01}{174.95\%}
& \paircell{6.559}{154.37\%}\\
\bottomrule
\end{tabular}
\end{table}

\begin{table}[!htbp]
\centering
\caption{Model-upgrade cost for one complete run over all six datasets. Each
\method{} row includes agent selection, one original test-set baseline, and
one post-upgrade test-set evaluation. FullSwap upgrades and evaluates every agent and
therefore has no DeepSeek-V3.2 selection-stage cost.}
\label{tab:model-upgrade-cost}
\scriptsize
\setlength{\tabcolsep}{2pt}
\resizebox{\textwidth}{!}{%
\begin{tabular}{llcccccc}
\toprule
Target & Method
& \shortstack{Qwen3-8B\\requests} & \shortstack{Qwen3-8B\\tokens (M)}
& \shortstack{DeepSeek-V3.2\\(selection)\\requests} & \shortstack{DeepSeek-V3.2\\(selection)\\tokens (M)}
& \shortstack{Target model\\(upgrade) requests} & \shortstack{Target model\\(upgrade) tokens (M)}\\
\midrule
Qwen3-32B & \textbf{IF-MAP} & \paircell{91{,}748}{100.00\%}
& \paircell{151.102}{100.00\%} & \paircell{794}{100.00\%}
& \paircell{4.028}{100.00\%} & \paircell{1{,}557}{100.00\%}
& \paircell{1.541}{100.00\%}\\
Qwen3-32B & FullSwap & \paircell{12{,}553}{13.68\%}
& \paircell{15.710}{10.40\%} & \paircell{0}{0.00\%}
& \paircell{0}{0.00\%} & \paircell{4{,}066}{261.14\%}
& \paircell{4.389}{284.82\%}\\
\addlinespace[2pt]
Qwen3-235B-A22B & \textbf{IF-MAP} & \paircell{91{,}748}{100.00\%}
& \paircell{151.102}{100.00\%} & \paircell{794}{100.00\%}
& \paircell{4.028}{100.00\%} & \paircell{1{,}534}{100.00\%}
& \paircell{1.601}{100.00\%}\\
Qwen3-235B-A22B & FullSwap & \paircell{12{,}518}{13.64\%}
& \paircell{15.730}{10.41\%} & \paircell{0}{0.00\%}
& \paircell{0}{0.00\%} & \paircell{4{,}085}{266.30\%}
& \paircell{4.481}{279.89\%}\\
\addlinespace[2pt]
Qwen3.7-Plus & \textbf{IF-MAP} & \paircell{91{,}748}{100.00\%}
& \paircell{151.102}{100.00\%} & \paircell{794}{100.00\%}
& \paircell{4.028}{100.00\%} & \paircell{1{,}524}{100.00\%}
& \paircell{1.781}{100.00\%}\\
Qwen3.7-Plus & FullSwap & \paircell{12{,}774}{13.92\%}
& \paircell{16.030}{10.61\%} & \paircell{0}{0.00\%}
& \paircell{0}{0.00\%} & \paircell{4{,}134}{271.26\%}
& \paircell{4.923}{276.42\%}\\
\addlinespace[2pt]
DeepSeek-V3.2 & \textbf{IF-MAP} & \paircell{91{,}748}{100.00\%}
& \paircell{151.102}{100.00\%} & \paircell{794}{100.00\%}
& \paircell{4.028}{100.00\%} & \paircell{1{,}522}{100.00\%}
& \paircell{1.939}{100.00\%}\\
DeepSeek-V3.2 & FullSwap & \paircell{12{,}871}{14.03\%}
& \paircell{16.210}{10.73\%} & \paircell{0}{0.00\%}
& \paircell{0}{0.00\%} & \paircell{4{,}601.2}{302.31\%}
& \paircell{5.331}{274.93\%}\\
\addlinespace[2pt]
GPT-5 nano & \textbf{IF-MAP} & \paircell{91{,}748}{100.00\%}
& \paircell{151.102}{100.00\%} & \paircell{794}{100.00\%}
& \paircell{4.028}{100.00\%} & \paircell{1{,}526}{100.00\%}
& \paircell{3.000}{100.00\%}\\
GPT-5 nano & FullSwap & \paircell{12{,}646}{13.78\%}
& \paircell{15.330}{10.15\%} & \paircell{0}{0.00\%}
& \paircell{0}{0.00\%} & \paircell{4{,}825.8}{316.24\%}
& \paircell{8.330}{277.67\%}\\
\bottomrule
\end{tabular}%
}
\end{table}

\FloatBarrier
\section{Artifact provenance and reproducibility}
\label{app:provenance}

All reported performance values are produced by the protocol specified in
this appendix. The experimental records contain the three prioritization
runs, five intervention runs, and the gain matrix used for each table.

The following materials and evaluations support a complete artifact release
and broader replication.
\begin{enumerate}[leftmargin=*,itemsep=3pt]
\item Example IDs, generated prompt texts, perturbation omissions, response
caches by repeat, and the gain matrix used for each table.
\item Per-call query, token, latency, and monetary accounting for the
prioritization methods and downstream interventions.
\item Evaluations on independently constructed workflows with different
interaction graphs and additional intervention families.
\item A public artifact at \url{https://github.com/s1412/IF-MAS} containing
the prompt snapshots, dependencies, evaluation commands, and applicable
licenses.
\end{enumerate}

Future topology studies could include code generation with
execution feedback, question decomposition with evidence checking, and
independent mathematical solution verification. Workflows with distinct
interaction graphs would test whether a node-response score remains useful
when information routing and correction opportunities change.

\end{document}